\documentclass[conference]{IEEEtran}
\IEEEoverridecommandlockouts
\usepackage{cite}
\usepackage{times}
\usepackage{soul}
\usepackage[small]{caption}
\usepackage{graphicx}
\usepackage{amsmath,amssymb,amsthm,bm}
\usepackage{algorithm}
\usepackage{algorithmic}
\usepackage{multirow}
\usepackage{bigstrut}
\usepackage{subfigure}
\usepackage{makecell, rotating}
\usepackage[utf8]{inputenc} 
\usepackage[T1]{fontenc}    
\usepackage{url}            
\usepackage{booktabs}       
\usepackage{amsfonts}       
\usepackage{nicefrac}       
\usepackage{microtype}      
\usepackage{xcolor}         

\usepackage[hidelinks]{hyperref}
\usepackage{bbding}
 
\let\softmax\relax
\DeclareMathOperator\softmax{softmax}

\newcommand{\squeezeup}{\vspace{-2mm}}

\def\BibTeX{{\rm B\kern-.05em{\sc i\kern-.025em b}\kern-.08em
    T\kern-.1667em\lower.7ex\hbox{E}\kern-.125emX}}
\begin{document}

\title{
    Self-Supervised Adversarial Example Detection by Disentangled Representation
}


\author{
Zhaoxi Zhang$^1$, 
Leo Yu Zhang$^{2}$, 
Xufei Zheng$^{1}$\Envelope, 
Jinyu Tian$^3$,  
Jiantao Zhou$^3$ \\
$^1$School of Computer and Information Science, Southwest University, China\\
$^2$School of Information Technology, Deakin University, Australia\\
$^3$Department of Computer and Information Science, University of Macau, Macau\\
\Envelope \ Corresponding Author: X. Zheng (zxufei@swu.edu.cn)
}

\maketitle

\begin{abstract}
Deep learning models are known to be vulnerable to adversarial examples that are elaborately designed for malicious purposes and are imperceptible to the human perceptual system. 
Autoencoder, when trained solely over benign examples, has been widely used for (self-supervised) adversarial detection based on the assumption that adversarial examples yield larger reconstruction errors. 
However, because lacking adversarial examples in its training and the too strong generalization ability of autoencoder, this assumption does not always hold true in practice. 
To alleviate this problem, we explore how to detect adversarial examples with disentangled label/semantic features under the autoencoder structure. 
Specifically, we propose Disentangled Representation-based Reconstruction (DRR). 
In DRR, we train an autoencoder over both correctly paired label/semantic features and incorrectly paired label/semantic features to reconstruct benign and counterexamples.
This mimics the behavior of adversarial examples and can reduce the unnecessary generalization ability of autoencoder.
We compare our method with the state-of-the-art self-supervised detection methods under different adversarial attacks and different victim models, and it exhibits better performance in various metrics (area under the ROC curve, true positive rate, and true negative rate) for most attack settings. 
Though DRR is initially designed for visual tasks only, we demonstrate that it can be easily extended for natural language tasks as well.
Notably, different from other autoencoder-based detectors, our method can provide resistance to the adaptive adversary. 
%
\end{abstract}

\section{Introduction}
\label{Sec:intro}
In 2013, the seminal work \cite{Szegedy2013} reported that, during model test time, deep neural networks can be easily fooled by adversarial attacks that add tiny perturbations to inputs. 
Since then, adversarial attacks and defenses have drawn significant research attention  \cite{Goodfellow2014,Moosavi-Dezfooli2016,Carlini2017b,Madry2017,Roth2019,Grosse2017, Hendrycks2016,Hu2022ProtectingFP,Hu2021AdvHashST,Zhang2022EvaluatingMI}.
On the one hand, attackers are persistently developing new strategies to construct adversarial examples; on the other hand, defenders are struggling to cope with all existing and forthcoming attacks \cite{Carlini2017}.

Most of the existing defense methods \cite{SID,Roth2019,Grosse2017,Lee2018,Hendrycks2016} are trained with supervision, and these methods work well when defending against adversarial attacks they were originally trained for.
However, it is widely regarded that supervised methods cannot generalize well to adversarial examples from (existing) unseen attacks, let alone examples from new attacks. 

Self-supervised based defense, in comparison with supervised defense, requires only benign examples for its training. 
As a typical example, the works in \cite{Meng2017,Chen2020} utilize the encoder of an autoencoder (AE) to draw the manifold of benign examples and then the decoder network for reconstruction, as shown in Fig.~\ref{fig_ae_vs_drr}(a). 
Since the manifold is learnt from benign examples only and the AE is trained to minimize Reconstruction Errors (REs) for benign examples, thus the encoding of adversarial examples will likely be out-of-distribution and the associated REs will be larger.

It is soon realized that this is not always true because AE has very strong generalization ability \cite{Gong2019}: examples with various kinds of small perturbations can be reconstructed with small RE. This is desirable if the perturbed examples are benign. 
However, adversarial examples are just specific perturbed versions of benign examples, and the malicious perturbations can also be made very small in many attacks (i.e., now their encodings will reside in the light gray area of Fig.~\ref{fig_ae_vs_drr}(a)). 
When this happens, all REs are mixed, and it leads to a high false negative or false positive rate (FNR/FPR) during detection. 
%
To refine the volume of the manifold drawn by AE and reduce its unnecessary generalization ability on adversarial examples, there exist a number of variants \cite{Vacanti2020,Wojcik2020,Qin2019,Schott2019,Yang2021ClassDisentanglementAA}, as will be reviewed in detail in Sec.~\ref{Sec:DeAdv}.


As a better solution, we propose a self-supervised disentangled representation-based reconstruction (DRR) method to detect adversarial examples. 
DRR possesses the advantage of supervised defense, even though it does not have access to any adversarial examples in training. %
This is achieved through mimicking the behavior of adversarial examples by encoding and decoding a special class of examples (counterexamples in this work), which is the reconstruction of the correct semantic feature and the incorrect label feature from one example.
The rationale is based on the very fact of adversarial examples: \textbf{they cause misclassification (i.e., wrong label) without changing semantics (contained in its benign counterparts).}

With this observation, we make use of the victim model and an auxiliary encoder to obtain disentangled representations: {class-dependent and class-independent features (i.e., label feature and semantic feature)}, of images.
After disentanglement, we then train a decoder to reconstruct benign examples by combining label/semantic features from the same benign image, as well as counterexamples by combining the original semantic feature and the post-processed incorrect label feature. 
For detection, as shown in Fig.~\ref{fig_ae_vs_drr}(b), DRR will faithfully disentangle any benign image into paired label/semantic features and its adversarial counterpart into unpaired label/semantic features, whose associated REs are significantly different as desired.

This paper makes the following contributions: 
\begin{itemize}
\item We use disentangled representation for adversarial detection, which makes it possible for the detector to mimic the behavior of adversarial examples in the self-supervised framework. 
\item We design DRR via an AE structure, but it reduces the unnecessary generalization capability of AE on adversarial examples. 
\item We achieve state-of-the-art adversarial detection performance on MNIST, SVHN, CIFAR-10 and CIFAR-100 in most cases. Specifically, DRR is the first of its kind that can offer protection against adaptive adversarial attacks. And we also demonstrate that DRR is effective for defending against adversarial textual attacks.
\end{itemize}

\section{Background and Related Works}
\label{Sec:relatedwork}

\subsection{Constructing Adversarial Examples}
The existence of adversarial examples in deep neural networks is first pointed out by \cite{Szegedy2013}, who find maliciously designed imperceptible perturbations can fool deep models to misclassify. 
Let $\mathsf{F}(\cdot)$ be a general neural model and $\mathsf{F}_z(\cdot)$ be the layers before $\softmax$ of the model, then evaluating the test example $x$ is simply a $\softmax$ classification over the logits $z = \mathsf{F}_z(x)$, i.e., 
$y = \mathsf{F}(x)  =\softmax(z)$. 
The (untargeted) adversarial example $x_{\textnormal{adv}}$ derived from $x$ satisfies
\begin{IEEEeqnarray}{rCl}
        &&\mathsf{F}(x_{\textnormal{adv}}) \neq   \mathsf{F}(x), 
        \nonumber \\
        && x_{\textnormal{adv}} = x+\delta_{\textnormal{adv}}  \textnormal{~and~}  \left\| \delta_{\textnormal{adv}} \right\| <  \epsilon,
        \label{Eq:advnorm}
\end{IEEEeqnarray}
where $\delta_{\textnormal{adv}}$ is the adversarial perturbation, $\epsilon$ is the maximum magnitude of $\delta_{\textnormal{adv}}$ under certain norm compliance. 
Commonly used norms include $L_2$ and $L_{\inf}$, and we also focus on detecting adversarial examples bounded by them. 

In the current literature, the Fast Gradient Sign Method (FGSM) is the first widely used method in generating adversarial examples \cite{Goodfellow2014}. 
Projected Gradient Descent (PGD) \cite{Madry2017} improves FGSM by iterating the building block of it according to different criteria to find the optimized perturbation. 
DeepFool \cite{Moosavi-Dezfooli2016} and CW \cite{Carlini2017b} do not directly rely on gradient but instead optimize (the norm of) $\delta_{\textnormal{adv}}$, which is usually more stealthy than FGSM and its variants.

Another line of research for constructing adversarial examples is called adaptive attacks \cite{carlini2017adversarial}. 
Under an adaptive attack, the attacker also has white-box access to possible defense mechanisms, and his goal is to find an optimal $\delta_{\textnormal{adv}}$ that solves Eq.~(\ref{Eq:advnorm}) and circumvents the defense mechanisms simultaneously. 
All the attacks above can have their adaptive versions by considering different defenses.

\begin{figure}[t]
    \centering
    \subfigure[]{\includegraphics[height=3cm]{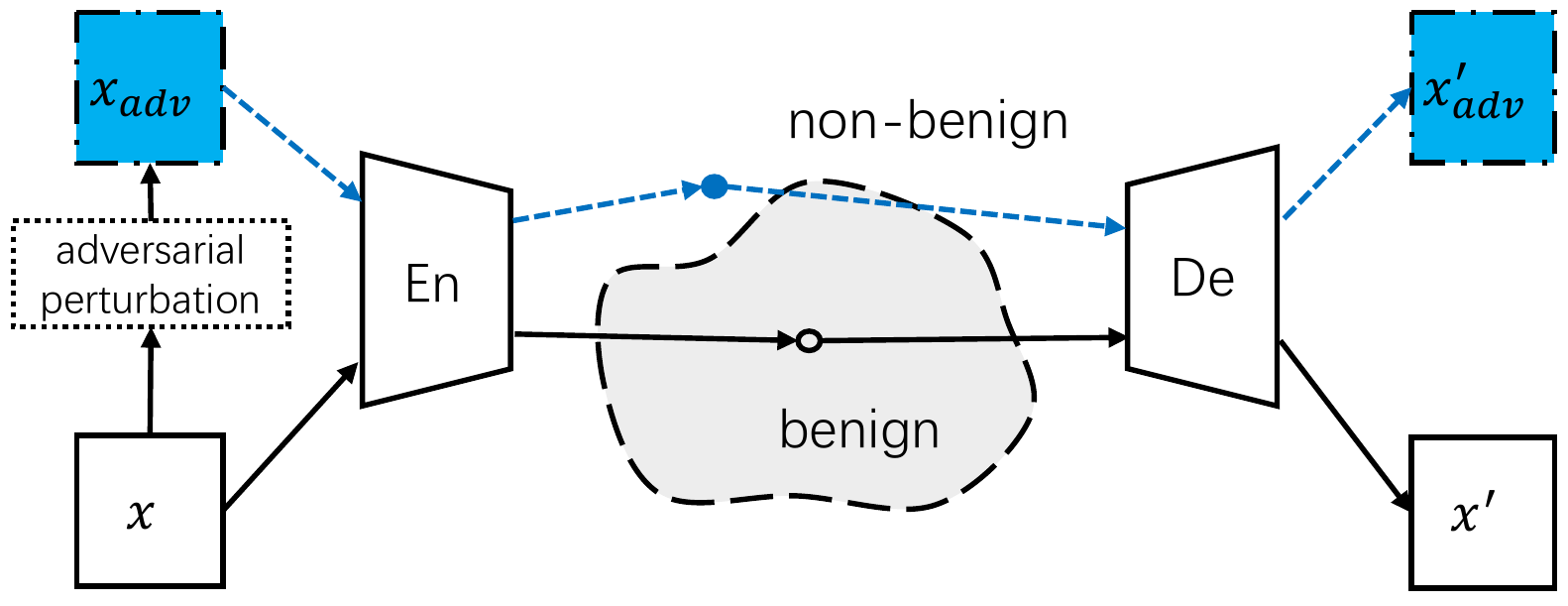}}
    \subfigure[]{\includegraphics[height=3cm]{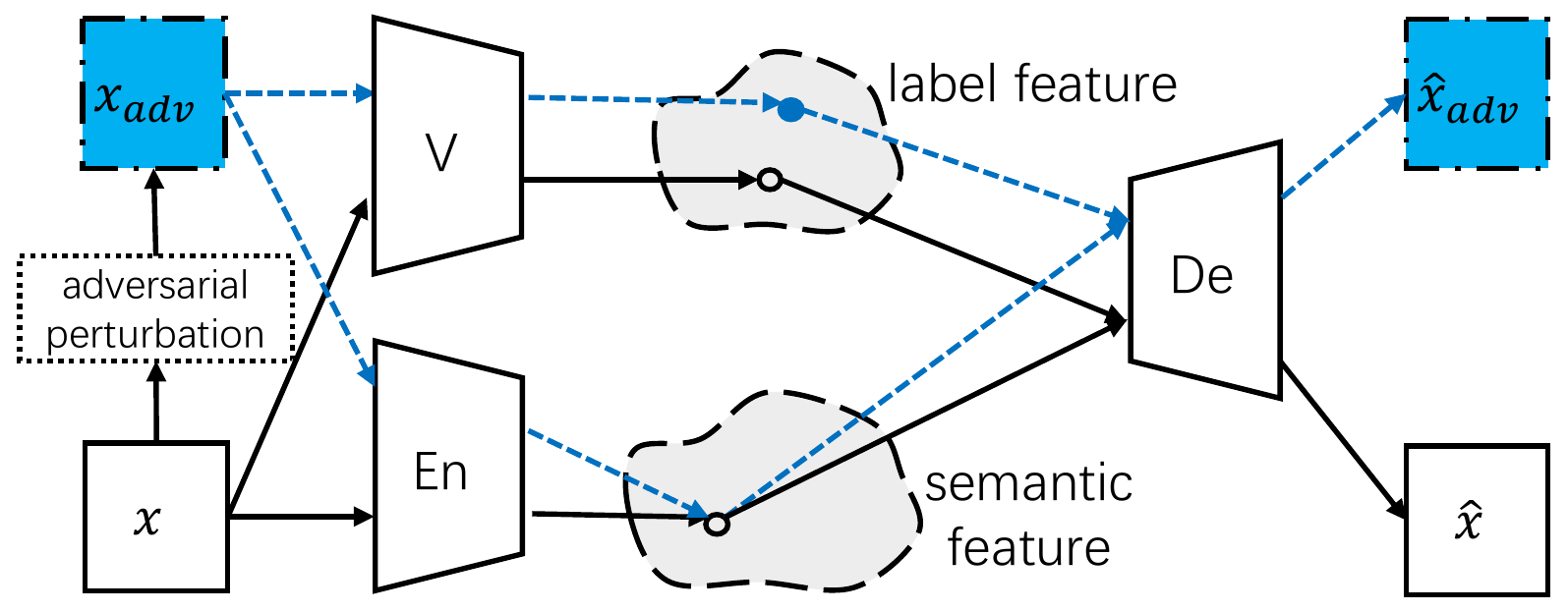}}
    \caption{
        Different detectors. 
        Here, $\mathsf{En}$ stands for the encoder, $\mathsf{De}$ the decoder, $\mathsf{V}$ the victim model. 
        (a) AE based method draws a manifold (the light gray area) of benign examples; 
        (b) DRR employs two disentangled features: label features which represent images' labels (vulnerable to adversarial perturbation) and semantic features (robust to adversarial perturbation). When decoding label/semantic features, benign examples can be reconstructed faithfully but adversarial examples cannot.
        }
    \label{fig_ae_vs_drr}
\end{figure}

\subsection{Detecting Adversarial Examples}
\label{Sec:DeAdv}
Adversarial detection is an effective way to prevent adversarial examples, and it can be classified as supervised and unsupervised methods, considering whether adversarial examples are needed to build a detector.
For supervised detectors, their detecting capability depends on how to capture the differences between adversarial and benign examples. 
Techniques range from studying statistical properties \cite{Grosse2017,Lee2018,Roth2019}, training traditional machine learning classifiers \cite{Lu2017,Feinman2017} and deep classifiers \cite{Wojcik2020,SID,Ma2018CharacterizingAS}. 
It is widely accepted that supervised methods cannot generalize well to adversarial examples produced by unseen attacks. 
For example, a supervised detector trained with PGD adversarial examples will likely fail to detect examples produced by DeepFool. 
However, it is also known that supervised detectors are robust to adaptive attacks, i.e., a detector trained with PGD can detect examples produced by the adaptive PGD attack \cite{SID,Ma2018CharacterizingAS}.

For unsupervised detectors, their detecting capability depends on how to embed and represent benign examples to a different manifold (other than the natural spatiotemporal domain) such that adversarial perturbations will be magnified with embedding (without even seeing them at all). 
The seminal work MagNet \cite{Meng2017} uses an autoencoder to draw the manifold for benign examples, which implicitly assumes that the distance between the adversarial and benign examples is large in this embedding manifold.

However, the embedding manifold induced by AE is not always desirable for detection since the autoencoder has a too strong generalization capability \cite{Gong2019}.
Further to MagNet, the work \cite{Vacanti2020} trains a variant of the autoencoder by adding logits of the victim model into the loss function to refine the volume of the embedded manifold.
The work \cite{Wojcik2020} proposes to directly use parameters fixed victim model as the encoder, and the victim models' logits as high-level representations/embedding.
Different from the works above, which treat the manifold of all benign examples as a whole, \cite{Schott2019,Qin2019,Yang2021ClassDisentanglementAA} propose a class-conditional model to embed and reconstruct benign examples for even better refining the (embedding) manifold.
However, it is reported in \cite{Qin2019} that this method still only reacts to relatively large (adversarial) perturbations over simple datasets. 
Compared with supervised detectors, self-supervised detectors may generalize better to unseen attacks but they are vulnerable to adaptive attacks \cite{Meng2017,Gong2019,Wojcik2020}. 
For example, it is shown in \cite{carlini2017magnet} that the detector of \cite{Meng2017} fails to resist adaptive attacks at all.

\subsection{Disentangled Representation}
\label{subsec:disrep}
Disentangled representation is firstly advocated by InfoGAN in \cite{Chen2016}, which encourages the learning of interpretable and meaningful representations of inputs for manipulating specific features. 
The work \cite{sang2020deaan} disentangles speaker-related representation to synthesis voices of different speakers and the work \cite{tran2017disentangled} uses disentangled representation to address the pose discrepancy problem among face images.

{From the view of information, the training of a neural model is to minimize the mutual information of input and output \cite{shwartz2017opening}, so semantic (contained in the input) and label (contained in the output) features of an image sample are made independent via training and they are disentangled by nature in any neural model.}
From the view of adversarial learning, it is acknowledged that there are many features in an image, and not all features are equally easy to be manipulated in adversarial attack \cite{Ilyas2019}. Specifically, the class an image belongs to (i.e., the label), which does not depend on semantic feature it has, is a concrete example. 
In this concern, we {employ the disentangled semantic feature and label feature to build DRR}.
Specifically, semantic feature can be easily manipulated by the attackers (through adversarial perturbation) but the perturbed semantic feature is similar to the original version (even after embedding if the perturbation is tiny). 
In contrast, label feature cannot be directly manipulated by attackers, but they are fragile to adversarial perturbation and robust to natural perturbation of the semantic feature. 


\begin{figure}[t]
    \centering 
    \includegraphics[height=3.5cm]{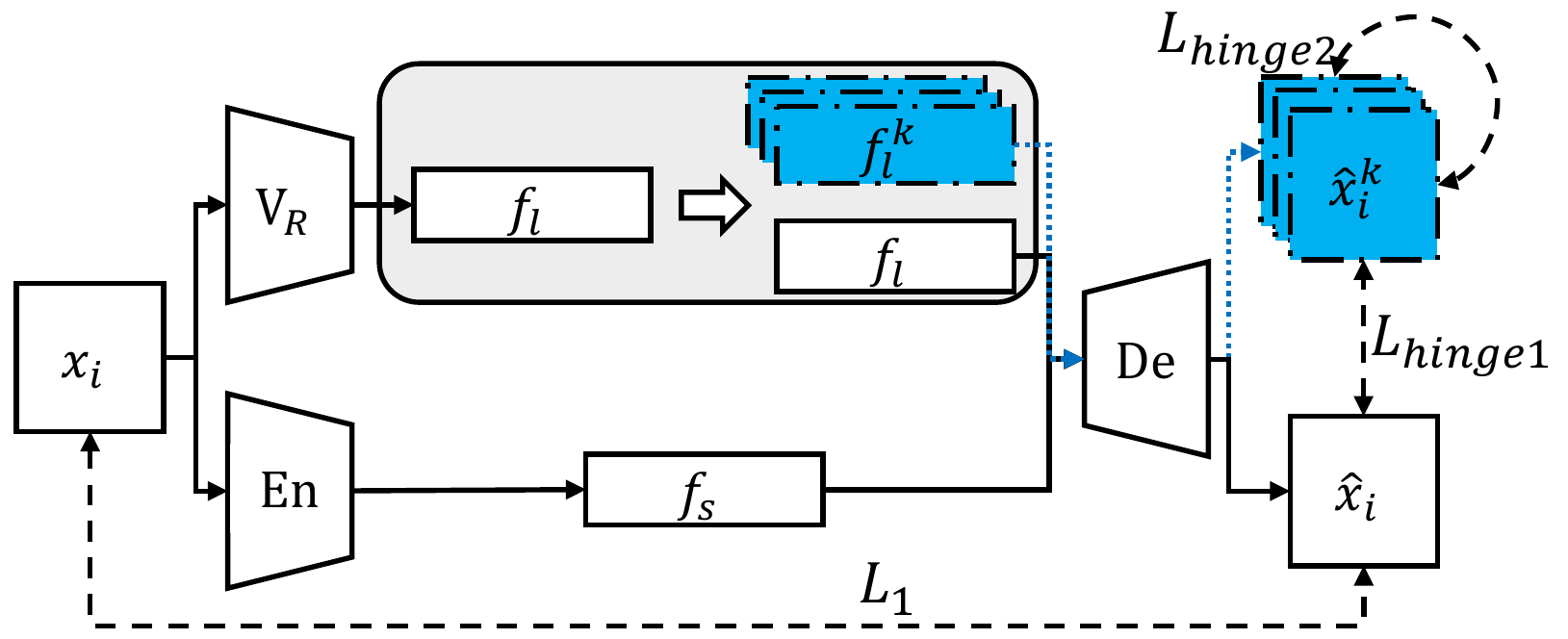}
    \caption{Overview of DRR.}
    \label{Fig:DRRtraining}
\squeezeup
\end{figure}

\section{DRR for Adversarial Detection}
\label{Sec:Methods}

\begin{figure*}[t]
    \centering 
    \subfigure[MNIST]{
        \parbox[b]{.3\textwidth}{
            \centering
            \includegraphics[height=0.9cm]{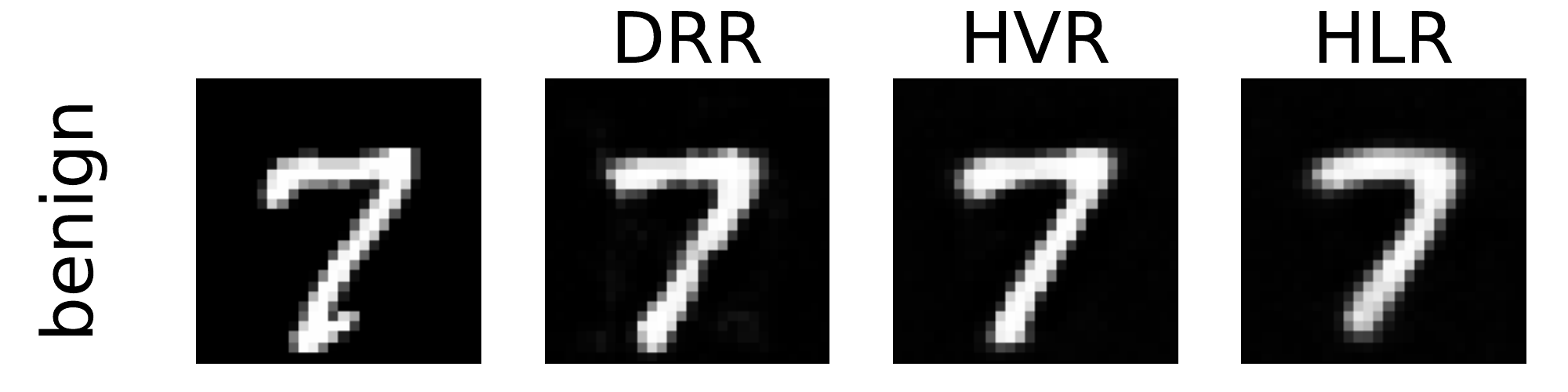}
            \includegraphics[height=0.9cm]{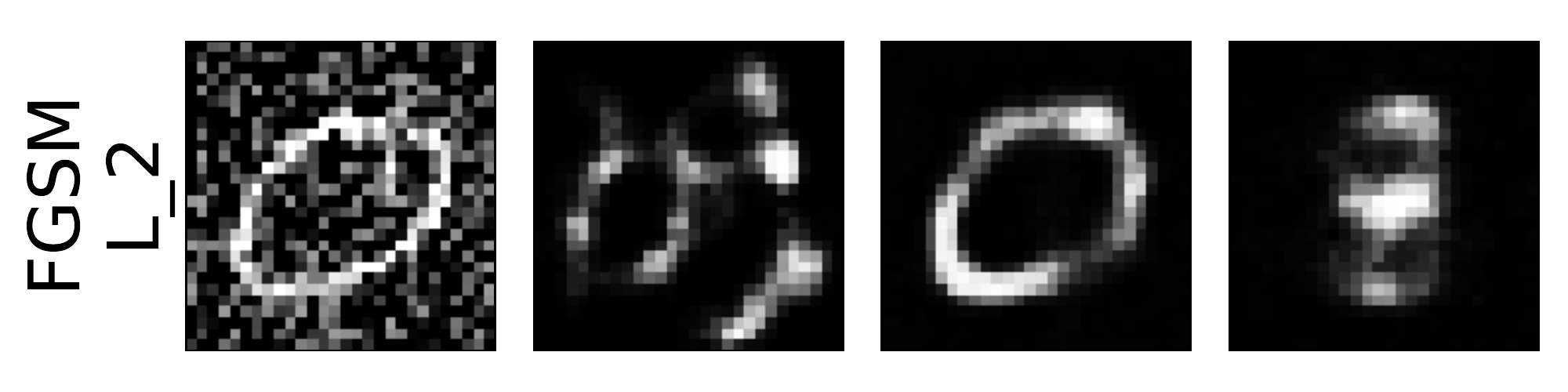}
            \includegraphics[height=0.9cm]{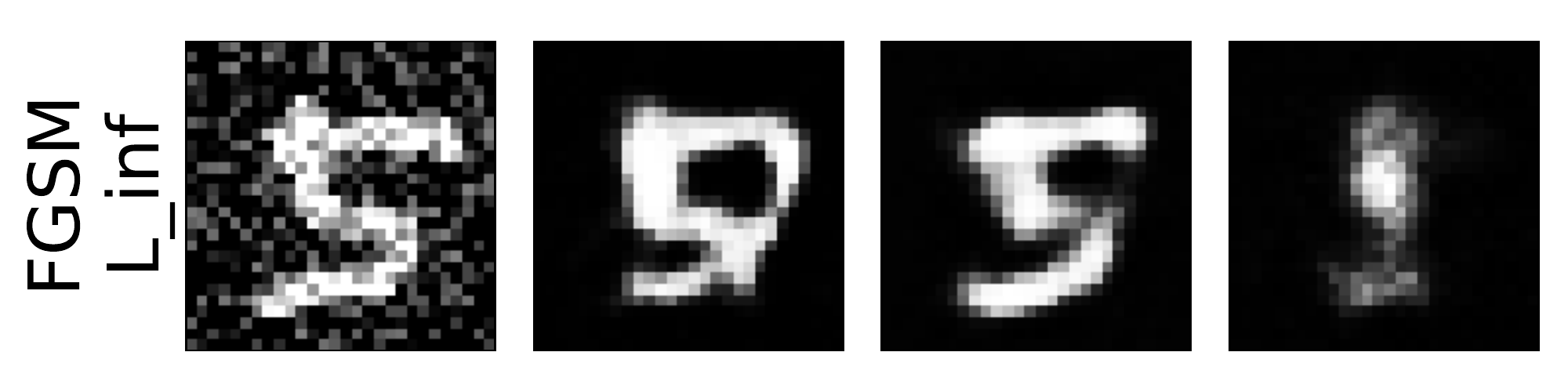}
            \includegraphics[height=0.9cm]{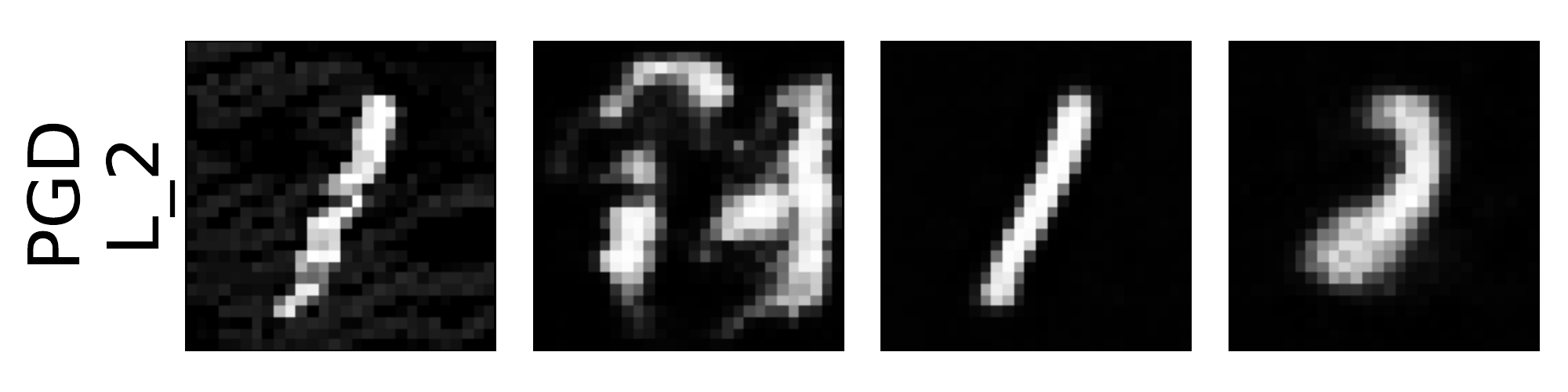}
            \includegraphics[height=0.9cm]{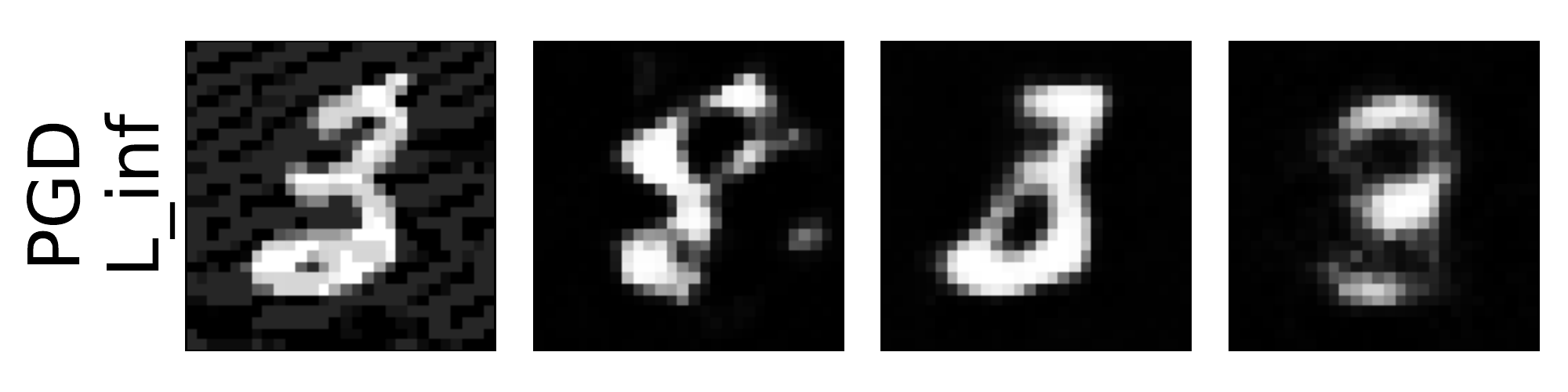}
        }
    }\quad
    \subfigure[SVHN]{
        \parbox[b]{.3\textwidth}{
            \centering
            \includegraphics[height=0.9cm]{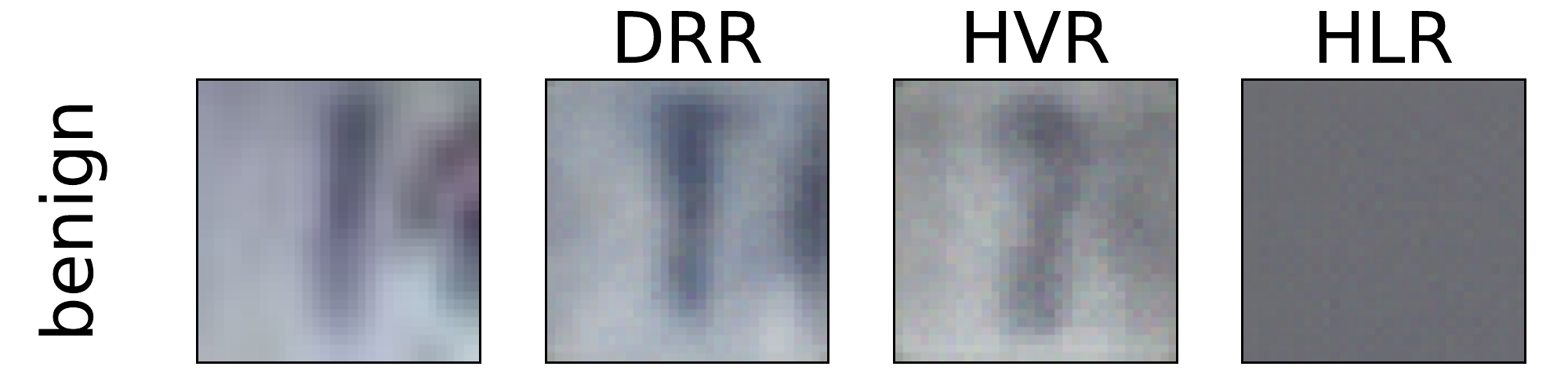}
            \includegraphics[height=0.9cm]{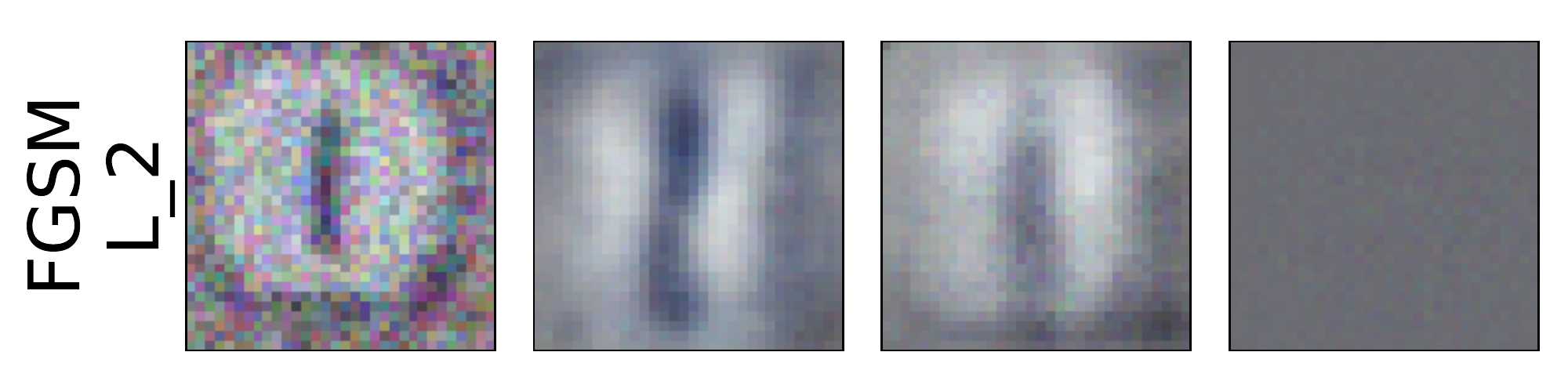}
            \includegraphics[height=0.9cm]{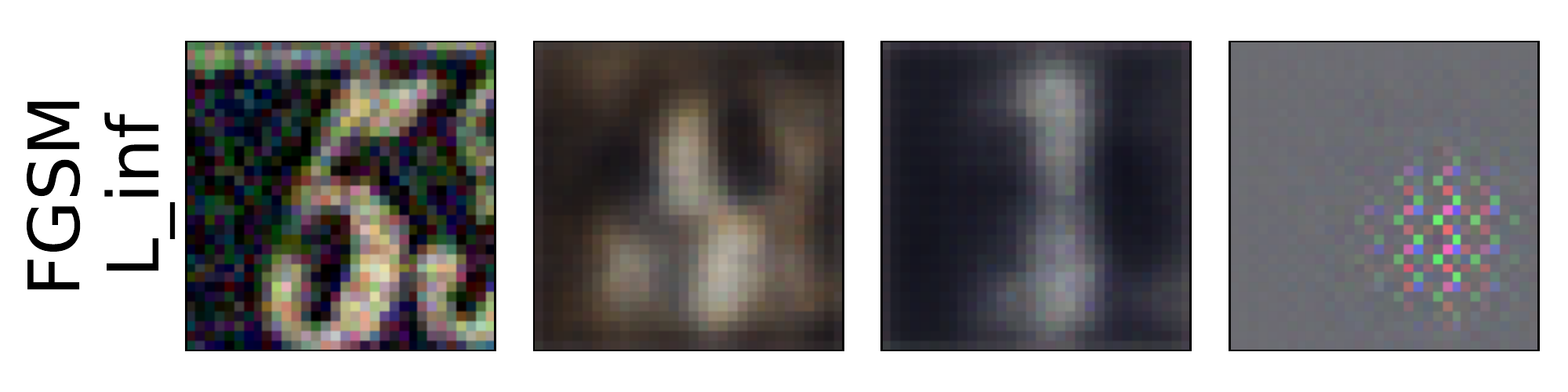}
            \includegraphics[height=0.9cm]{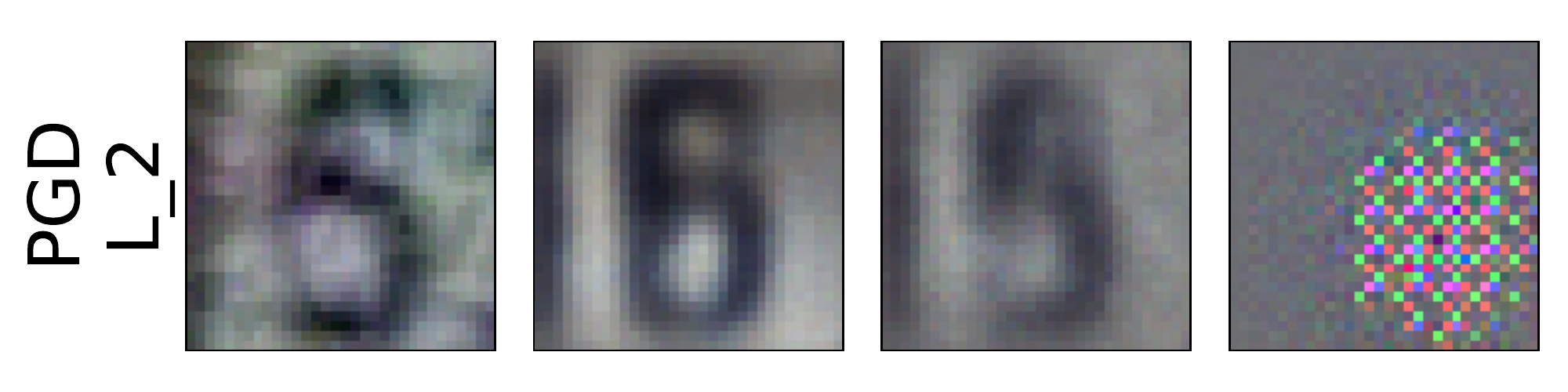}
            \includegraphics[height=0.9cm]{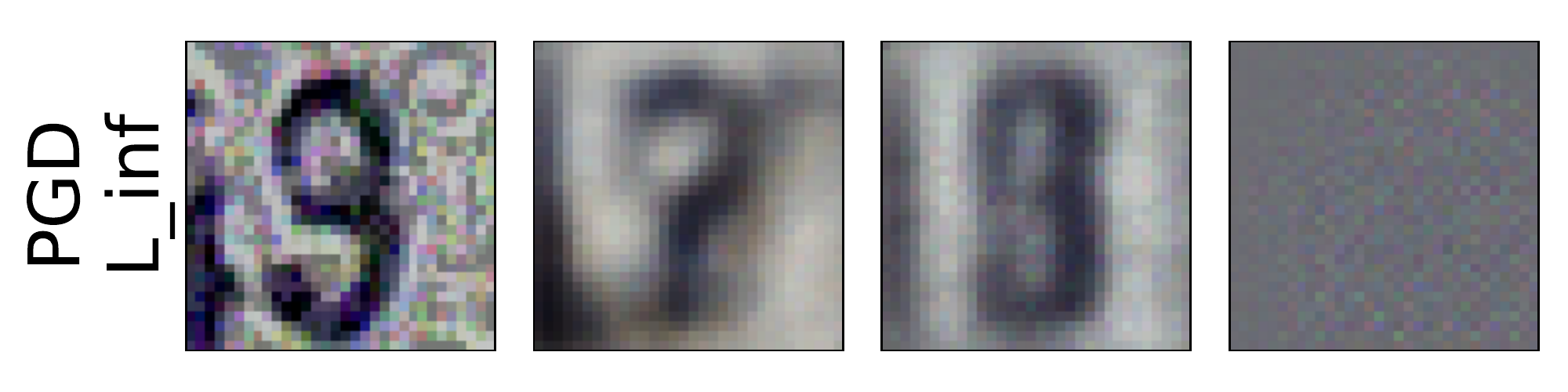}
        }
    }\quad
    \subfigure[CIFAR10]{
        \parbox[b]{.3\textwidth}{
            \centering
            \includegraphics[height=0.9cm]{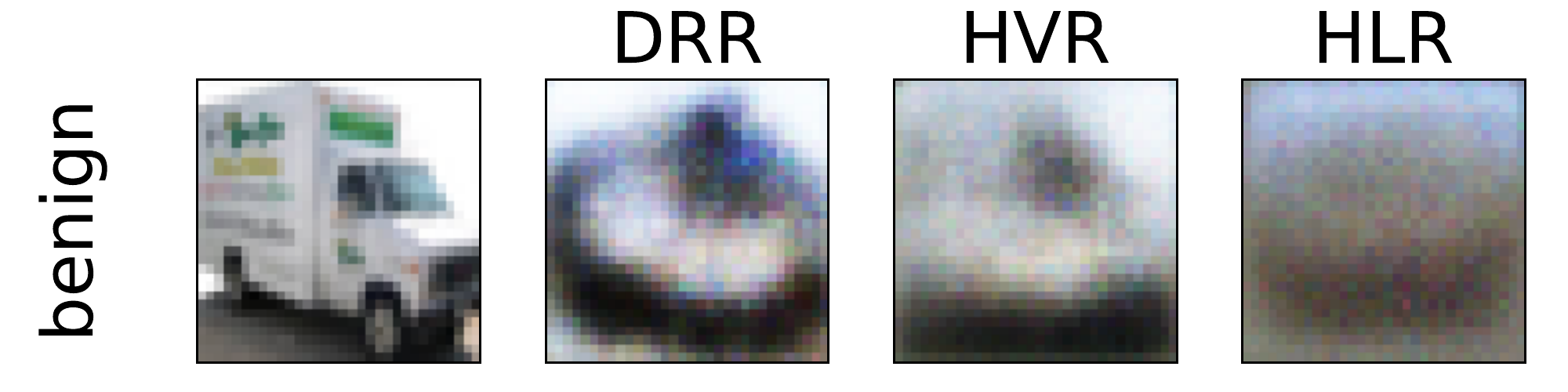}
            \includegraphics[height=0.9cm]{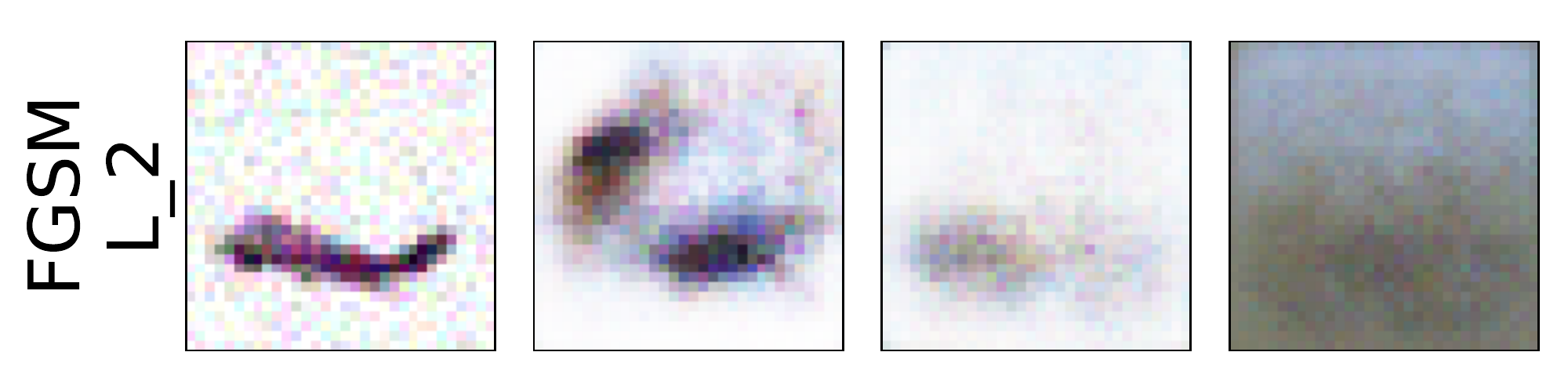}
            \includegraphics[height=0.9cm]{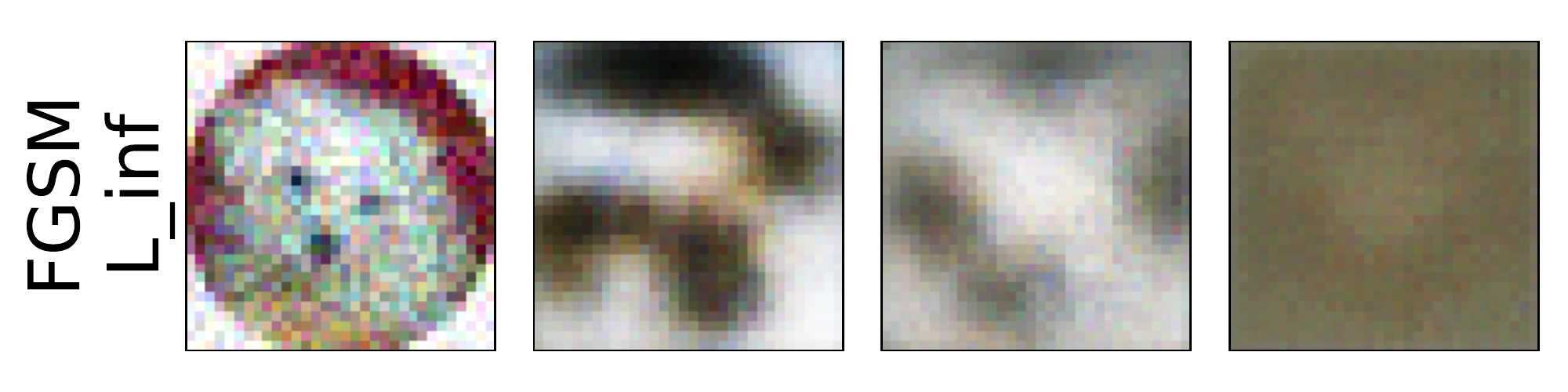}
            \includegraphics[height=0.9cm]{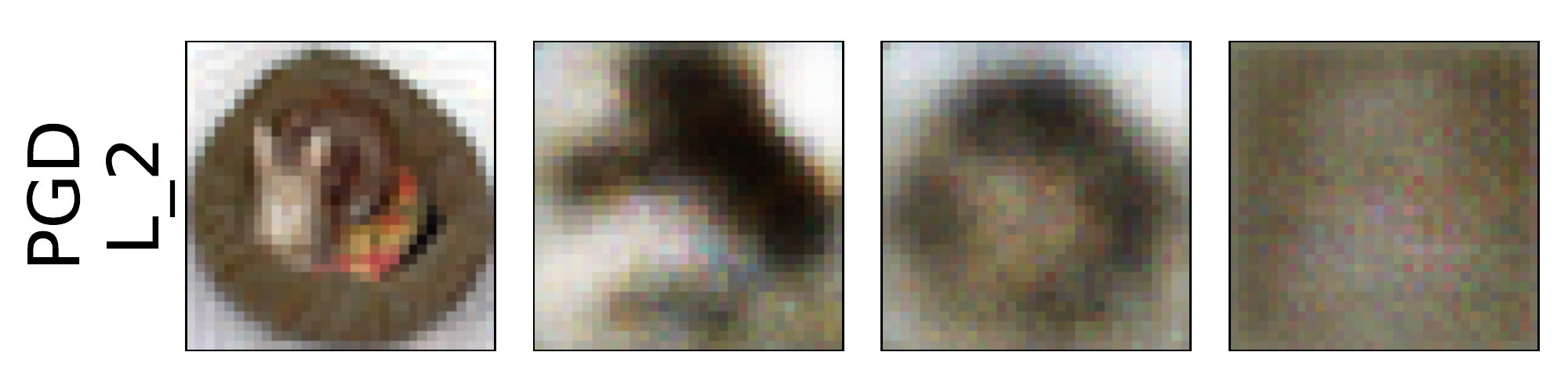}
            \includegraphics[height=0.9cm]{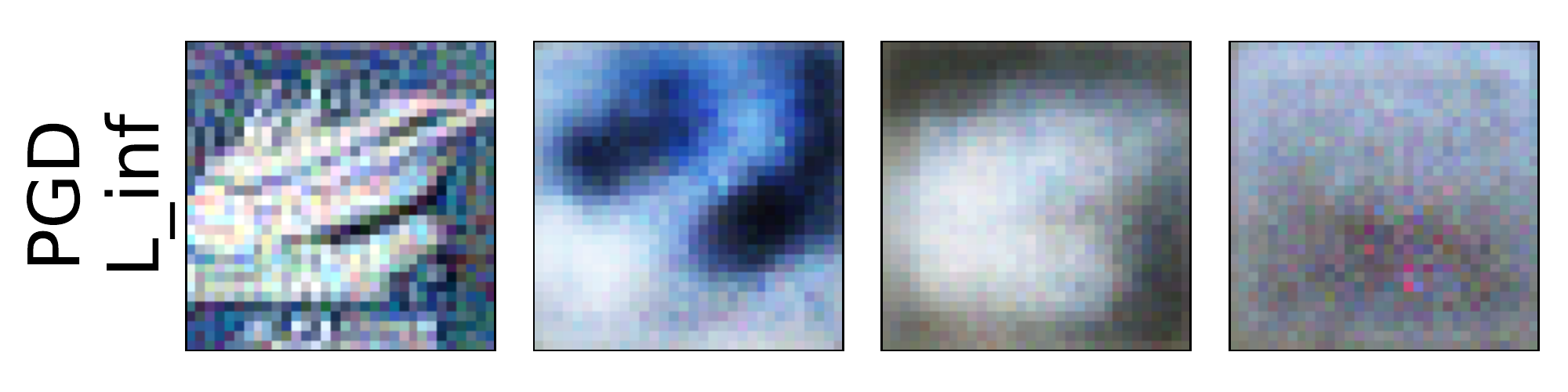}
        }
    }
    \centering 
    \caption{Reconstructed images of each AE-based detector.}
    \label{fig:visualoutput}
\end{figure*}

\subsection{Training of DRR}
\label{Sec:train_drr}
From a high-level-of-view, as depicted in Fig.~\ref{Fig:DRRtraining}, assisted by the parameter-fixed victim model $\mathsf{V}$, DRR first extracts {label and semantic} features from benign examples $x_i$ to train the encoder $\mathsf{En}$ and decoder $\mathsf{De}$. 
Then DRR permutes the label feature of $x_i$ and combines it with the semantic feature of $x_i$ to decode counterexamples. 
The details of how to design the loss functions of $\mathsf{En}$ and $\mathsf{De}$ of DRR are presented in detail below. 

First and foremost, {we examine how to obtain the desired semantic feature $f_{s}$ and label feature $f_{l}$ of $x_i$ for detection purpose.}
For $f_{l}$, we choose to rescale the logits the victim $\mathsf{V}$, i.e., 
\begin{IEEEeqnarray}{rCl}
        y &=& \mathsf{V}(x_i) = \softmax(\mathsf{V}_z(x_i)) ,\\
        f_{l} &=& \mathsf{V}_R(x_i) = R_S(\mathsf{V}_z(x_i)),
\end{IEEEeqnarray}
where $\mathsf{V}_R(x_i)$ denotes the rescaled logits and the rescaling function is given by
\begin{IEEEeqnarray}{rCl}
        R_S(x) &=& \softmax(x/T),
        \label{Eq:rescale_func}
\end{IEEEeqnarray}
where  $T \geq 1$ is a temperature parameter. The choice of the label feature $f_{l}$ is two-folded: 1) logits is very close to the final one-hot encoding of $\mathsf{V}(x_i)$ and resacling with a temperature $T$ is commonly used for information distillation \cite{Hinton2015DistillingTK}; 2) no matter how the concrete adversarial example $x_{\textnormal{adv}}$ is constructed, $\mathsf{V}_R(x_{\textnormal{adv}})$ must be different enough from $\mathsf{V}_R(x)$ to induce the final erroneous classification. 

%

For the semantic feature $f_{s}$, we use a general encoder network $\mathsf{En}$ to derive it, i.e., $f_{s} = \mathsf{En}(x_i)$. 
This is because it is widely accepted that high-dimensional input $x_i$ resides in a low-dimensional manifold, and it is a commonly used method in the area of representation learning.

With $f_{l}$ and $f_{s}$ available, the decoder network $\mathsf{De}$ of DRR reconstructs $x_i$ as $\hat{x}_i = \mathsf{De} \left( f_{s}, f_{l} \right)$. The natural requirement for $\mathsf{En}$ and $\mathsf{De}$ is that, for a generic image $x_i$ from the benign dataset, the reconstructed version $\hat{x}_i$ should be similar to $x_i$. 
Thus, the associated loss is: 
\begin{IEEEeqnarray}{rCl}
    \textnormal{L}_1 &=& \mathbb{E}_{x_i} \mathsf{MAE}(x_i, \hat{x}_i),
    \label{Eq:loss_reconstruction}
\end{IEEEeqnarray}
where $\mathsf{MAE}$ is the mean absolute error and $\mathbb{E}$ the expectation. 



We then move to the training of DRR with counterexamples by using unpaired semantic and label features, as depicted by the dashed blue box in Fig.~\ref{Fig:DRRtraining}. 
As emphasized in Sec.~\ref{Sec:DeAdv}, the drawback of AE based detection is AE generalizes too well and the refinement of the manifold drawn by AE is not always effective, especially on attacks that directly optimize the norm of the adversarial perturbation.
The solution is now straightforward since we can mimic the behavior of adversarial examples by constructing counterexamples from unpaired semantic feature and label feature to better refine the manifold drawn by $\mathsf{En}$ and $\mathsf{De}$. 

To obtain counterexamples, we first randomly relocate the largest element in $f_{l}$ from index $i$ to $k$, i.e.,  
\begin{IEEEeqnarray}{rCl}
    f^k_{l_i}&=&\mathsf{R}_L(f_{l_i}, k), k\neq i.
\end{IEEEeqnarray}
We then decode the permuted label feature $f^k_{l_i}$ and the original semantic feature $f_{s_i}$ to obtain counterexamples by $\hat{x}^k_i = \mathsf{De} \left( f_{s_i}, f^k_{l_i} \right)$.

Recall that our original purpose is to refine the manifold drawn by the encoder and decoder, so the loss function here should satisfy the following two requirements: 
\begin{itemize}
    \item The decoded $\hat{x}^k_i$ should not converge to $\hat{x}_i$ (and thus $x_i$) because convergence indicates that counterexamples and benign examples are mixed, which is contrary to our original detection purpose;
    \item The decoded $\hat{x}^k_i$ should not be far away from $\hat{x}_i$, since too-far-away implies out-of-distribution and will cause overfitting.
\end{itemize}
Note that $\hat{x}^{k_1}_i$ and $\hat{x}^{k_2}_i$ ($k_1 \neq k_2$) are different counterexamples of $x_i$ and they should also obey the above requirements.
For these reasons, we propose to use a soft hinge function as the loss, i.e., 
\begin{IEEEeqnarray}{rCl}
    \textnormal{L}_{hinge1} &=& \mathbb{E}_{x_i} \left[ \max \Big(0, d-\min_{\substack{k_1, k_2 \in \Sigma,\\ k_1 \neq k_2}}\big( \mathsf{MAE}( \hat{x}^{k_1}_i, \hat{x}^{k_2}_i ) \big) \Big) \right],\nonumber \\ \\
    \textnormal{L}_{hinge2} &=& \mathbb{E}_{x_i} \left[ \max \Big(0, d-\min_{ \substack{k\in \Sigma,\\ k \neq i}} \big( \mathsf{MAE}( \hat{x}^k_i, x_i ) \big) \Big) \right], \\
    \textnormal{L}_{hinge}  &=& \textnormal{L}_{hinge1}+\textnormal{L}_{hinge2}, 
    \label{Eq:hinge_loss}
\end{IEEEeqnarray}
where $d$ is a hyperparameter used to control the farthest allowable distance between counter and benign examples, the set $\Sigma$ determines the number of counterexamples used for $x_i$. 
This soft hinge function (passively) meets the second requirements list above: its gradient equals $0$ when $\mathsf{MAE}( \hat{x}^{k}_i, \hat{x}_i ) > d$, {so when the loss meets its upper bound, it will not contribute gradient to the training process.} 
{The size of the set $\Sigma$ is also an important hyperparameter because a hard counterexample contributes more to the final performance than an easy counterexample. When the size of $\Sigma$ is too small, the chance of sampling hard counterexamples is low each time. But when the size of $\Sigma$ is too large, it will cause loss function very hard to converge.}

In summary, the final loss function to train the encoder $\mathsf{En}$, and the decoder $\mathsf{De}$ of DRR is
\begin{IEEEeqnarray}{C}
    \textnormal{Loss} = \lambda \textnormal{L}_{1} + \textnormal{L}_{hinge},
\end{IEEEeqnarray}
where $\lambda$ is used to control the relative importance. 


\subsection{Detecting with DRR}
For detection, as discussed earlier and shown in Fig.~\ref{fig_ae_vs_drr}(b), an incoming test example $x$ will be encoded and decoded as $\hat{x}$, then the RE $\left\| \hat{x}-x \right\|_2$ is compared to a threshold value. If the RE is larger, then it is considered as adversarial; otherwise, it is not. Ideally, this threshold should be universal (on a given dataset) for all possible attacks (even for new attacks). However, for existing detectors, the threshold value is related to the used attack methods. This makes it hard to determine a universal threshold for existing detectors and to compare with them. For this reason, we use {the metrics area under the ROC curve (AUC), true positive rate (TPR) and true negative rate (TNR)} for evaluation in Sec.~\ref{Sec:results}. But it is worth mentioning that it is easy to set a universal threshold for DRR for all the considered attacks.


\section{Experimental Results}
\label{Sec:results}
In this section, we assess the performance of DRR on $4$ datasets, MNIST, SVHN and CIFAR-10/CIFAR-100, by comparing it with other state-of-the-art self-supervised detectors \cite{Meng2017,Vacanti2020,Wojcik2020,Yang2021ClassDisentanglementAA} against the adversarial attacks FGSM, PGD, DeepFool (taken from Foolbox \cite{rauber2017foolbox}) and the adaptive PGD under differnt norms ($L_2$ and $L_{\inf}$). 
As a proof-of-concept, two representative networks, an $8$-layer CNN and VGG-16 \cite{Simonyan2014}, are used as the victim models. 

\begin{figure*}[t]
    \centering 
        \subfigure[FGSM $L_{\textnormal{inf}}$]{\includegraphics[height=3.0cm]
        {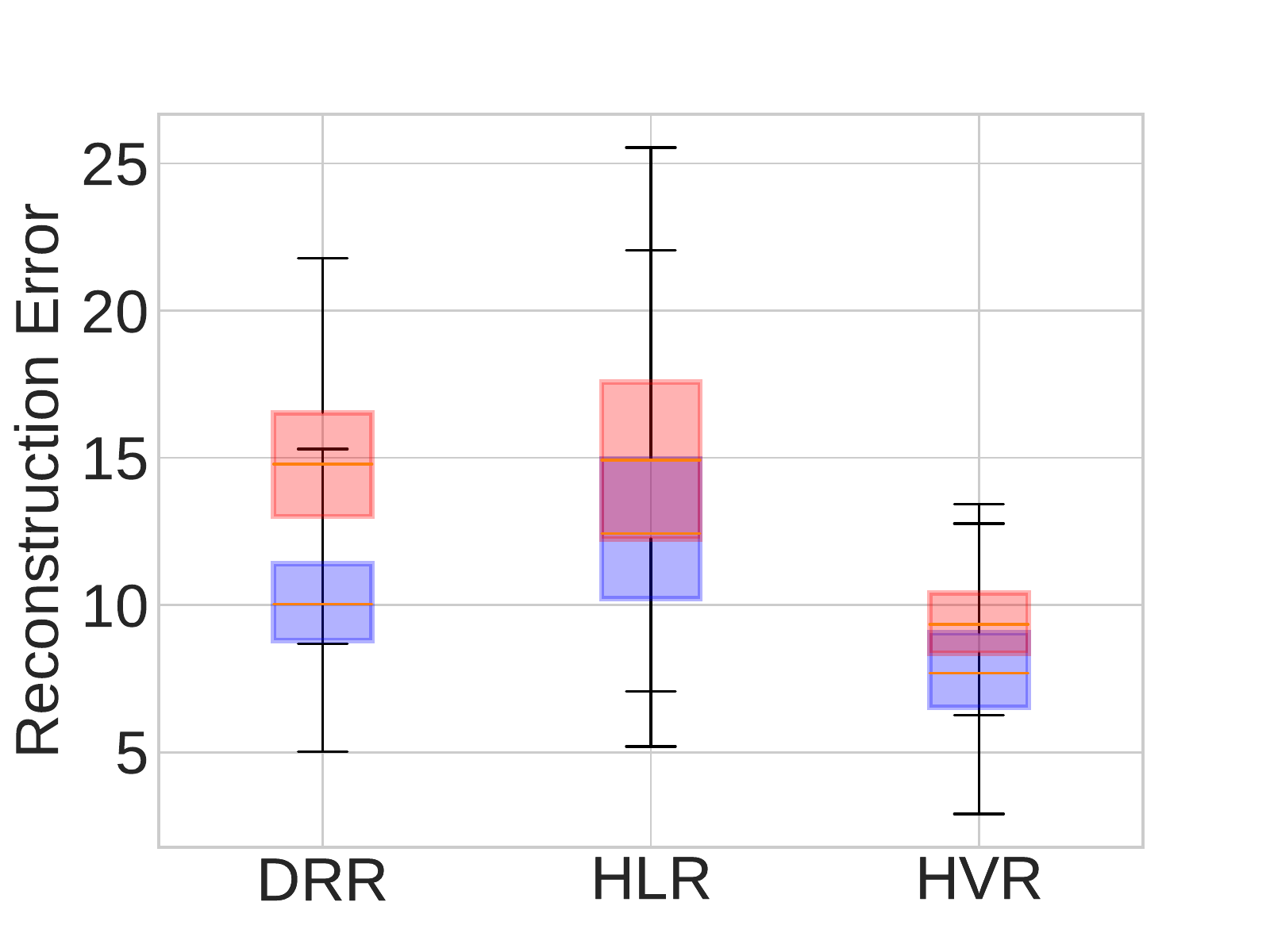}}
        \subfigure[PGD $L_{\textnormal{inf}}$]{\includegraphics[height=3.0cm]
        {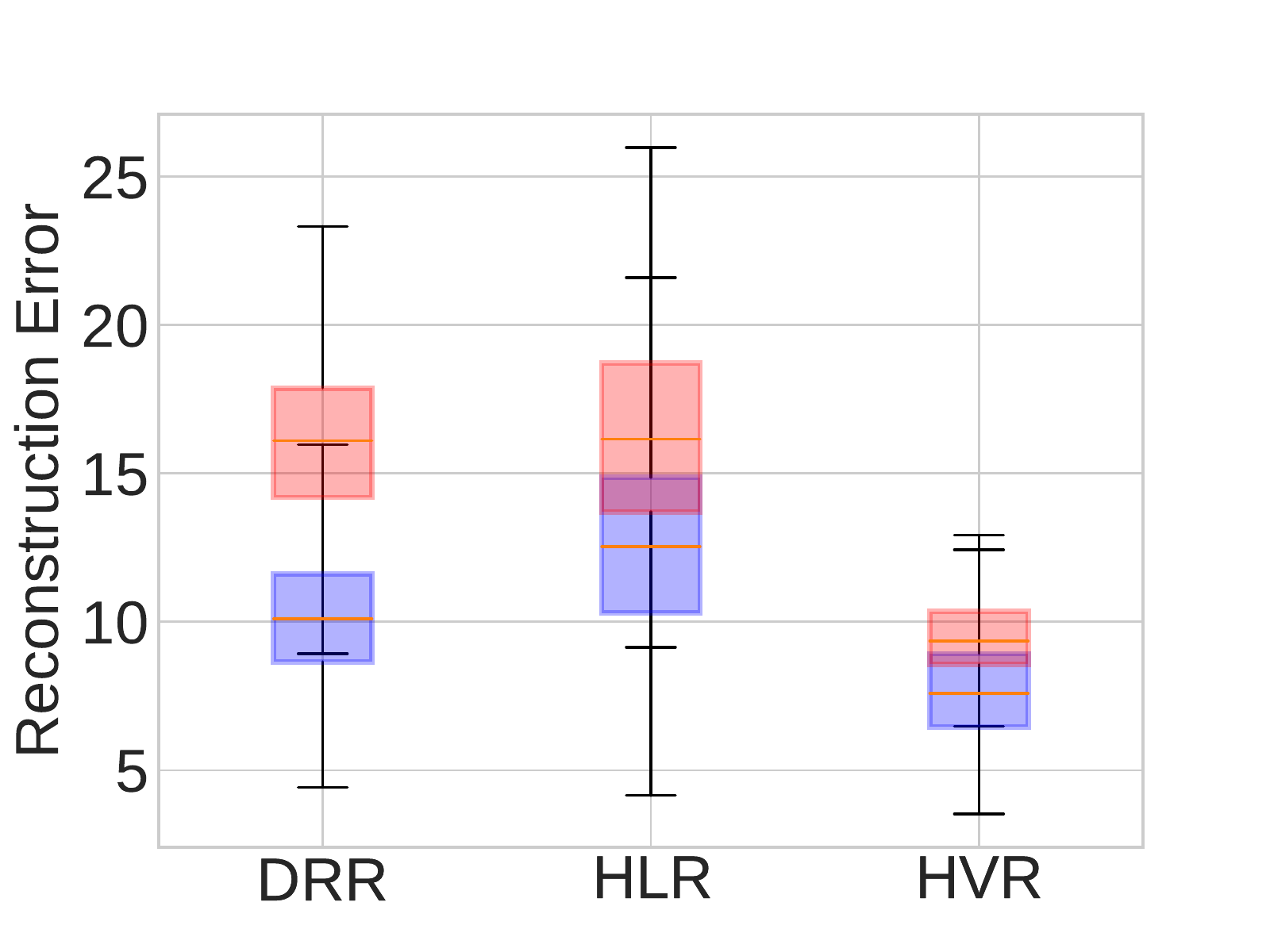}}
        \subfigure[FGSM $L_{\textnormal{inf}}$]{\includegraphics[height=3.0cm]
        {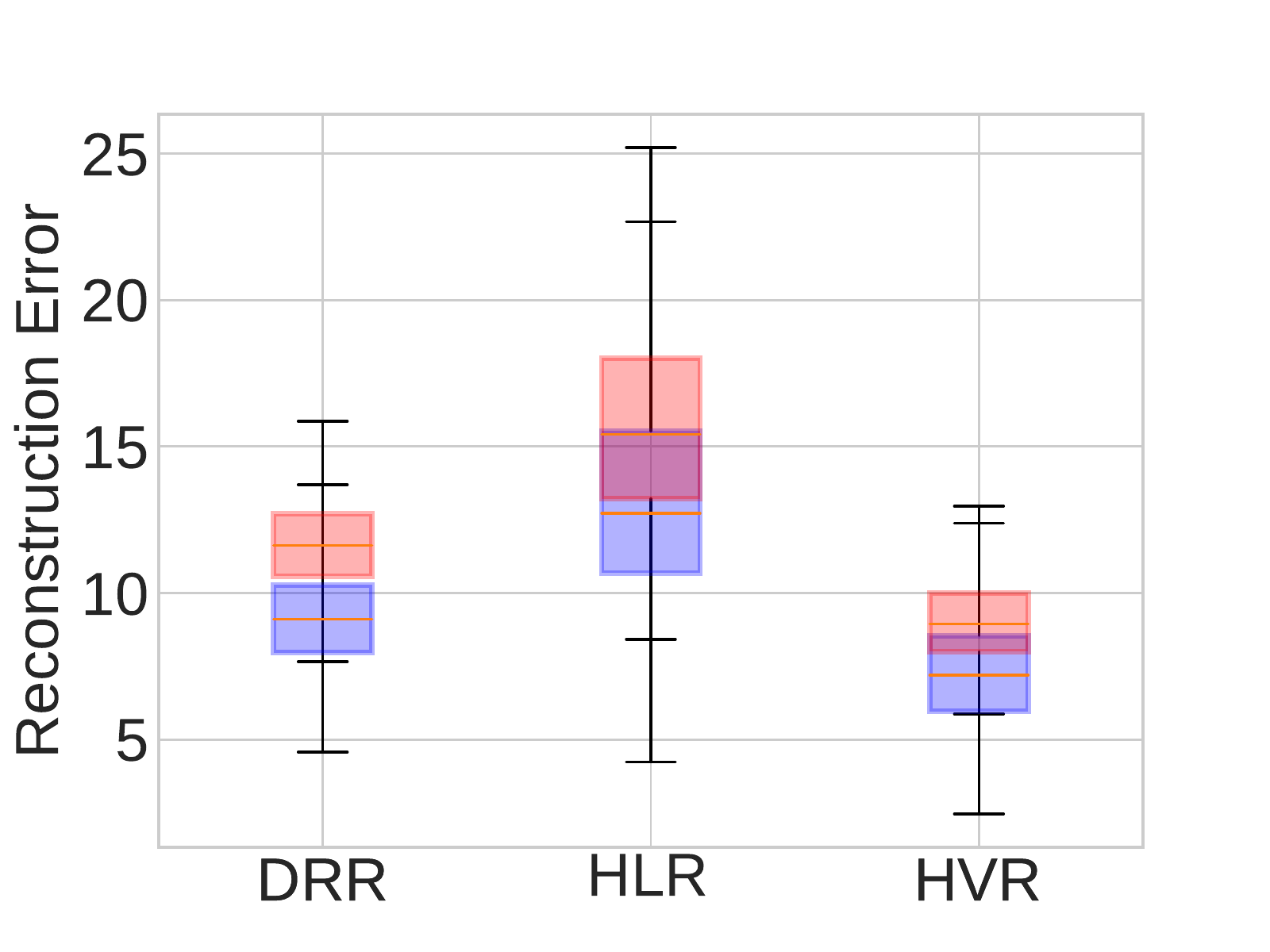}}
        \subfigure[PGD $L_{\textnormal{inf}}$]{\includegraphics[height=3.0cm]
        {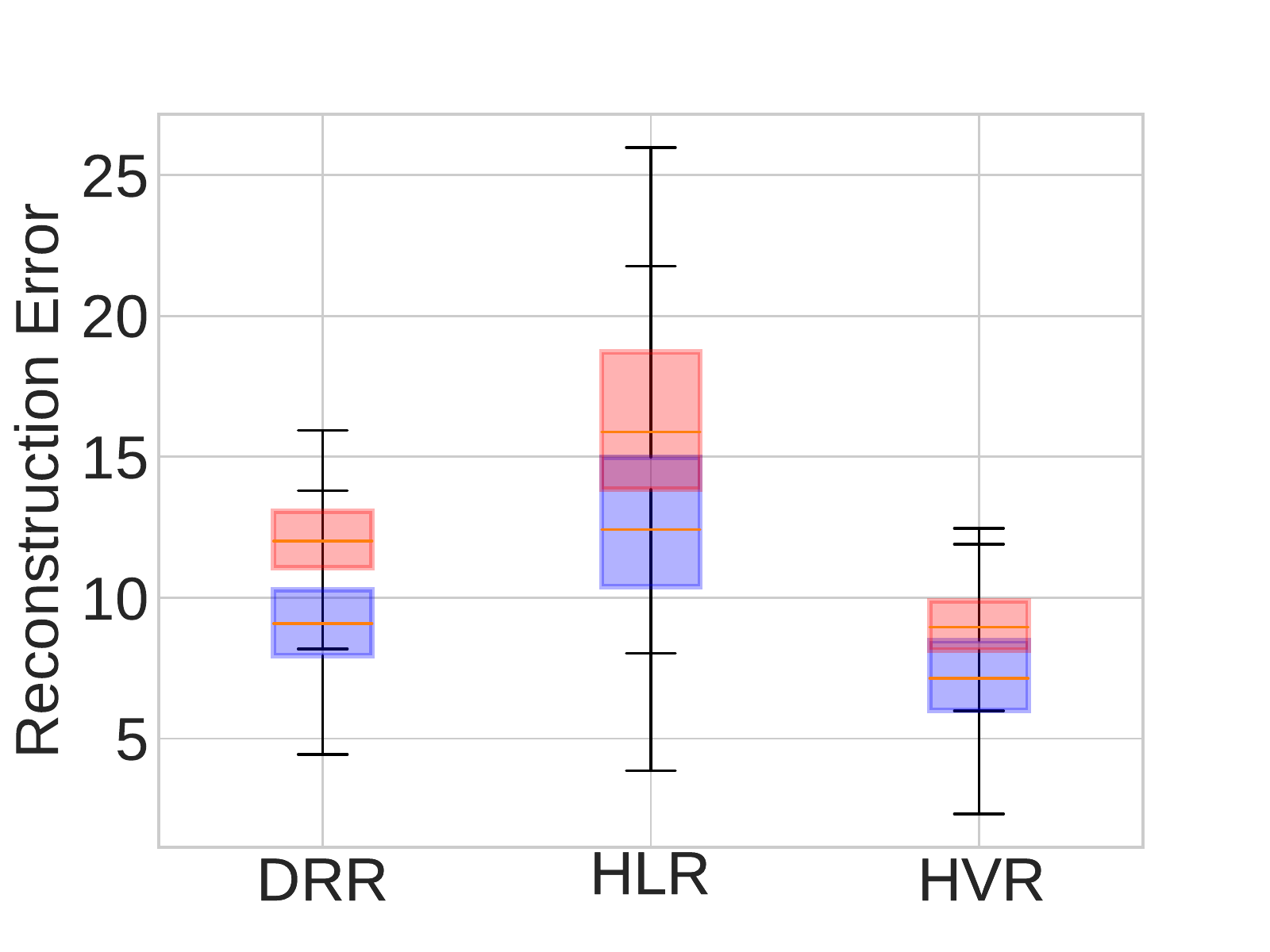}}
        \caption{
            Box-plot of the reconstruction errors: (a, b) CIFAR-10, and (c, d) CIFAR-100   (red box is for adversarial and blue box for benign). 
            }
        \label{fig_boxplot}
\end{figure*}

\subsection{Settings}
\noindent \textbf{DRR}: 
DRR consists of the encoder $\mathsf{En}$ and the decoder $\mathsf{De}$. 
It is suffice to say that $\mathsf{En}$ has the same architecture as the victim model except that its last layer is fully connected and contains $64$ neurons.
The architecture of the decoder $\mathsf{De}$ is simply a mirror of $\mathsf{En}$. 
Following the literature \cite{Ma2018CharacterizingAS,SID}, the hyperparameter $\lambda$ of the \textnormal{Loss} are determined by a binary search in $[10^{-2}, 10^2]$ 
($3$, $5$, $4$, $3$ for MNIST, SVHN, CIFAR-10, CIFAR-100). The hyperparameter $d$ is  determined empirically as $0.16$, $0.16$, $0.2$, $0.14$ for MNIST, SVHN, CIFAR-10, CIFAR-100, respectively.
{The size of the set $\Sigma$ is set to $20$ for CIFAR100, and $8$ for the other datasets.}

\noindent \textbf{Methods for comparison}: 
As mentioned above, three kinds of self-supervised detectors \cite{Meng2017, Vacanti2020, Wojcik2020, Yang2021ClassDisentanglementAA} are used for comparison. 
The structure of the encoder, the decoder for all these three methods are the same as of DRR.
These methods can be classified as hidden vector based reconstruction (HVR) and high-level representation based reconstruction (HLR). 
For case of HVR, the works in \cite{Meng2017, Vacanti2020, Yang2021ClassDisentanglementAA} are based on optimizing the pixel-level RE from the hidden vector outputted by the encoder. 
For case of HLR, the work in \cite{Wojcik2020} takes the logits from the parameter-fixed victim model as a high-level representation of examples for decoding. 

\subsection{Results Analyses}

\noindent \textbf{The baseline results}: 
{The victim VGG-16 is trained on SVHN, CIFAR-10, CIFAR-100 and the victim 8-layer CNN is trained on MNIST, these $4$ victim models achieve $0.961$, $0.869$, $0.682$, $0.993$ accuracy over benign test examples.}
For each test example, we generate its corresponding adversarial versions with $3$ attacks (FGSM, PGD, DeepFool) under $L_2$ and $L_{\inf}$ norm compliance ($\epsilon$ in Eq.~(\ref{Eq:advnorm})). 
Following the literature works \cite{Meng2017,Vacanti2020,Wojcik2020}, for FGSM and PGD attacks on MNIST, we set the budget for $L_{\inf}$ as $0.5$ and $0.15$, and the budget for $L_2$ as $10$ and $2$. 
For SVHN and CIFAR-100, we set the budget for $L_{\inf}$ as $0.15$ and $0.15$, and the budget for $L_2$ as $5$ and $2$. 
For CIFAR-10, we set the budget for $L_{\inf}$ as $0.15$ and $0.15$, and the budget for $L_2$ as $5$ and $1$. 
Note that DeepFool directly optimizes adversarial perturbation, so no budget setting is needed. 
After attack, the accuracy of $4$ victim models over adversarial examples all lower than $0.1$. 

\noindent \textbf{Visual inspection}: 
We first visually compare the reconstruction of the benign and adversarial examples for all the AE-based detectors, some examples are shown in Fig.~\ref{fig:visualoutput}. 
Inspecting cols. $2$ and $4$ of Fig.~\ref{fig:visualoutput}, it is clear that DRR and HLR generally have bigger REs over adversarial examples, which validates that logits (of the victim model) reduce undesirable generalization capability of the detectors. 

We further box-plot the REs of all benign and adversarial examples from CIFAR-10 and CIFAR-100 in Fig.~\ref{fig_boxplot}. 
It is clear from this figure, for the attacks on CIFAR-10 and CIFAR-100, the REs of DRR are clearly separable. 
The REs of HVR is also generally separable, but it has a higher FNR or FPR than that of DRR (depending on the concrete threshold of RE). And HLR performs the worst among all settings. 
This validates that DRR not only prevents undesirable generalization (large REs over adversarial examples), but also retains the merits of HVR-based methods (small REs over benign examples). 

\begin{table*}[t]
    \centering
    \caption{AUC/TPR of different detectors over different datasets. TPR is calculated when TNR=$0.9$.}
        \begin{tabular}{c|l|ccc}
            \toprule
            \multicolumn{1}{l|}{Dataset} & Attack & \multicolumn{1}{c}{HVR} & \multicolumn{1}{c}{HLR} & \multicolumn{1}{c}{DRR} \\
            \midrule
            \multirow{6}[2]{*}{MNIST} & FGSM $L_2$ & \textbf{0.999 / 1.000} & 0.985 / \textbf{1.000} & 0.998 / \textbf{1.000} \\
                  & FGSM $L_\infty$ & \textbf{0.998 / 1.000} & 0.967 / 0.954 & 0.995 / \textbf{1.000} \\
                  & PGD $L_2$ & 0.655 / 0.152 & 0.827 / 0.450 & \textbf{0.981 / 0.962} \\
                  & PGD $L_\infty$ & 0.804 / 0.349 & 0.856 / 0.495 & \textbf{0.983 / 0.972} \\
                  & DeepFool $L_2$ & 0.994 / 0.984 & 0.991 / 0.977 & \textbf{0.997 / 0.995} \\
                  & DeepFool $L_\infty$ & 0.997 / 0.994 & 0.996 / 0.993 & \textbf{0.998 / 1.000} \\
            \midrule
            \multirow{6}[2]{*}{SVHN} & FGSM $L_2$ & 0.919 / 0.629 & 0.636 / 0.173 & \textbf{0.955 / 0.930} \\
                  & FGSM $L_\infty$ & 0.915 / 0.662 & 0.620 / 0.129 & \textbf{0.948 / 0.893} \\
                  & PGD $L_2$ & 0.593 / 0.113 & 0.539 / 0.106 & \textbf{0.889 / 0.579} \\
                  & PGD $L_\infty$ & 0.918 / 0.677 & 0.643 / 0.125 & \textbf{0.985 / 0.986} \\
                  & DeepFool $L_2$ & 0.869 / 0.761 & 0.815 / 0.621 & \textbf{0.941 / 0.878} \\
                  & DeepFool $L_\infty$ & 0.827 / 0.627 & 0.704 / 0.343 & \textbf{0.910 / 0.742} \\
            \midrule
            \multirow{6}[2]{*}{$\substack{\mathrm{CIFAR}\\10}$} & FGSM $L_2$ & 0.757 / 0.308 & 0.673 / 0.249 & \textbf{0.911 / 0.758} \\
                  & FGSM $L_\infty$ & 0.761 / 0.288 & 0.678 / 0.222 & \textbf{0.904 / 0.744} \\
                  & PGD $L_2$ & 0.505 / 0.099 & 0.616 / 0.190 & \textbf{0.839 / 0.539} \\
                  & PGD $L_\infty$ & 0.778 / 0.326 & 0.767 / 0.360 & \textbf{0.942 / 0.851} \\
                  & DeepFool $L_2$ & 0.819 / 0.646 & 0.777 / 0.538 & \textbf{0.899 / 0.778} \\
                  & DeepFool $L_\infty$ & 0.749 / 0.519 & 0.677 / 0.377 & \textbf{0.848 / 0.703} \\
            \midrule
            \multirow{6}[2]{*}{$\substack{\mathrm{CIFAR}\\100}$} & FGSM $L_2$ & 0.750 / 0.287 & 0.717 / 0.246 & \textbf{0.844 / 0.485} \\
                  & FGSM $L_\infty$ & 0.754 / 0.292 & 0.713 / 0.227 & \textbf{0.859 / 0.535} \\
                  & PGD $L_2$ & 0.550 / 0.114 & 0.649 / 0.214 & \textbf{0.710 / 0.273} \\
                  & PGD $L_\infty$ & 0.772 / 0.255 & 0.745 / 0.257 & \textbf{0.894 / 0.632} \\
                  & DeepFool $L_2$ & 0.923 / 0.854 & 0.876 / 0.752 & \textbf{0.927 / 0.861} \\
                  & DeepFool $L_\infty$ & 0.861 / 0.714 & 0.804 / 0.590 & \textbf{0.881 / 0.743} \\
            \bottomrule
        \end{tabular}%
    \label{tab:auc_tpr}%
\end{table*}%

\noindent 
\textbf{AUC and TPR}: 
We quantitatively study the effectiveness of all detectors over the $4$ datasets by tabulating the results of AUC of ROC and TPR in Table~\ref{tab:auc_tpr}. 
From this table, it is clear that DRR consistently performs the best in terms of  true positive detection rate when keeping the true negative detection rate at $0.9$. 
Since all detectors aim for binary classification, the larger the size of the AUC of ROC, the better the detector. Ideally, AUC of ROC is $1$. 
For the SVHN, CIFAR-10 and CIFAR-100 datasets, the AUC of DRR is the largest over all different attacks. For MNIST, DRR outperforms other detectors for most cases with the exceptions when all detectors' AUC values are close to $1$. Since these exceptions occur for simple attack (FGSM) on simple dataset (MINST), we speculate all the detectors' performances are roughly the same on these cases. 


\section{Ablation Study}

We study the influences of $\textnormal{L}_{hinge}$ by considering $3$ Combined Representation Reconstruction (CRR) methods defined below:
\begin{IEEEeqnarray}{rCl}
    \mathrm{CRR0}: \textnormal{Loss} &=& \lambda \textnormal{L}_{1}, \\
    \mathrm{CRR1}: \textnormal{Loss} &=& \lambda \textnormal{L}_{1} + \textnormal{L}_{hinge1}, \\
    \mathrm{CRR2}: \textnormal{Loss} &=& \lambda \textnormal{L}_{1} + \textnormal{L}_{hinge2}.
    \label{Eq:ablation_hinge}
\end{IEEEeqnarray}
Except for the difference in Loss function, all CRR variants inherit the same settings from DRR. 
{And they share the same model architecture with DRR, which means they all use encoder $\mathsf{En}$ to generate semantic features and use Eq.~(\ref{Eq:rescale_func}) to get label features, then they reconstruct input data with the help of decoder $\mathsf{De}$. 
Both $\mathrm{CRR1}$ and $\mathrm{CRR2}$ have a part of $\textnormal{L}_{hinge}$, this enables use further evaluate the contribution of each part of $\textnormal{L}_{hinge}$. 
Note that HVR uses semantic feature as hidden state representation, while HLR uses logits (label feature) as hide state representation.}
So, from the view of architecture, $\mathrm{CRR0}$ can be seen as the combination of HLR and HVR.

Table~\ref{tab:ablation_hinge} lists the performance of these CRRs and the original DRR when detecting PGD attack with $L_\infty$ (recall that $\epsilon=0.15$) on CIFAR10.
It is clear from this table (and Table~\ref{tab:auc_tpr}) that the direct combination of HLR and HVR, i.e., CRR0, does not promote detection performance. 
Moreover, from this table, it is easy to see that $\textnormal{L}_{hinge1}$ is the main reason for promoting performance, and the standalone usage of $\textnormal{L}_{hinge2}$ harms detection. {We noted that without the help of $\textnormal{L}_{hinge1}$, $\textnormal{L}_{hinge2}$ is hard to converge in training.}
Only when combining $\textnormal{L}_{hinge1}$ and $\textnormal{L}_{hinge2}$ will the detection performance be boosted.

\begin{table}[h]
    \caption{AUC and TPR (TNR=$0.9$) for DRR and CRRs.}
    \label{tab:ablation_hinge}%
    \centering
      \begin{tabular}{l|rrrr}
      \toprule
            & \multicolumn{1}{l}{CRR0} & \multicolumn{1}{l}{CRR1} & \multicolumn{1}{l}{CRR2} & \multicolumn{1}{l}{DRR} \\
      \midrule
      AUC   & 0.769 & 0.885 & 0.726 & 0.942 \\
      TPR   & 0.278 & 0.732 & 0.211 & 0.851 \\
      \bottomrule
      \end{tabular}%
    \captionsetup{type=table}
\end{table}

\section{Defensing Adaptive Adversarial Attack}
To further evaluate the performance of DRR, we assume that the attacker can not only access the victim model but also knows all the details of the detector. Under this adaptive assumption, the attacker's goal is to fool both the victim model and the detector. 
Following the most-widely used adaptive attack strategy \cite{SID,carlini2017magnet,carlini2017adversarial}, the attacker now aims to solve: 
\begin{IEEEeqnarray}{rCl}
        & \min_{x_{\textnormal{adv}}} \ ~~ \alpha\textnormal{L}_{\textnormal{RE}}(x_{\textnormal{adv}}) - \textnormal{L}_{\textnormal{CE}}(x_{\textnormal{adv}},y) \nonumber \\
        & \textnormal{s.t.~~} \left\| x_{\textnormal{adv}}-x \right\|_\textnormal{inf} <\epsilon, 
        \label{Eq:adpattack}
\end{IEEEeqnarray}
where $\textnormal{L}_{\textnormal{RE}}$ and $\textnormal{L}_{\textnormal{CE}}$ are respectively the attack's loss function for the detector (i.e., reconstruction error) and for the victim model (i.e., cross entropy), and $y=\mathsf{F}(x)$ is the true label of $x$ and $\alpha$ is the parameter to control the relative importance of the two loss functions. 
It is worth mentioning that the detectors \cite{Meng2017,Vacanti2020,Wojcik2020} in comparison fail to resist the adaptive attack defined above\footnote{We note that \cite{Vacanti2020} considered a weaker adaptive attack strategy (maximize the cross entropy after AE reconstruction but not directly optimize reconstruction error) and HVR is robust to the weaker adaptive notion.}. 
The reason is, for other AE-based detectors, no matter high-level representation (i.e., logits) is used or not, they fail to capture true features of adversarial examples by certain. In contrast, the AE in DRR mimics the behavior of adversarial examples via constructing counterexamples with disentangled representation. 

\begin{figure*}[t]
    
    \centering 
        \centering 
        \subfigure[acc. after attack]{\includegraphics[height=3.0cm]{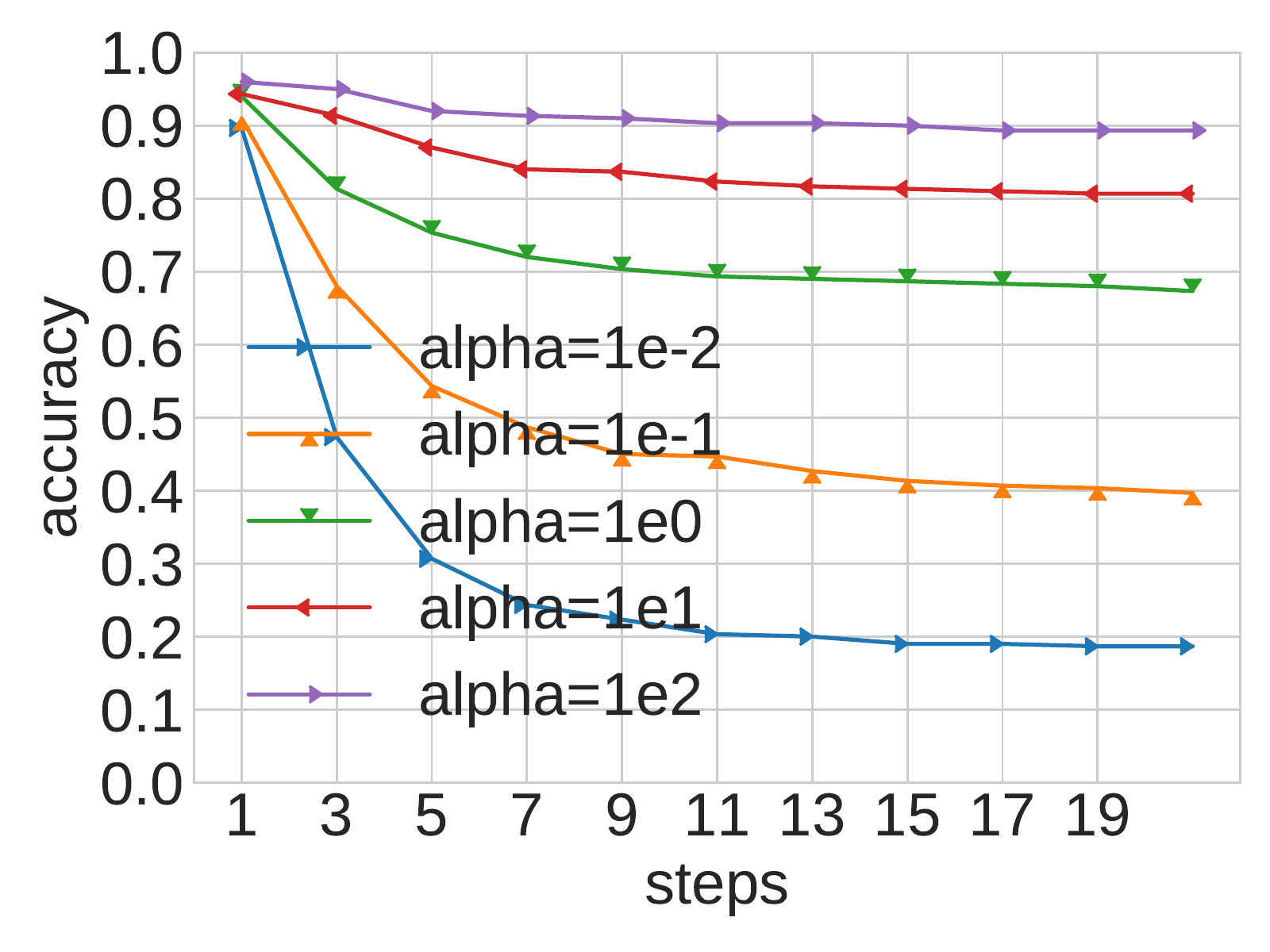}}
        \subfigure[AUC of detection]{\includegraphics[height=3.0cm]{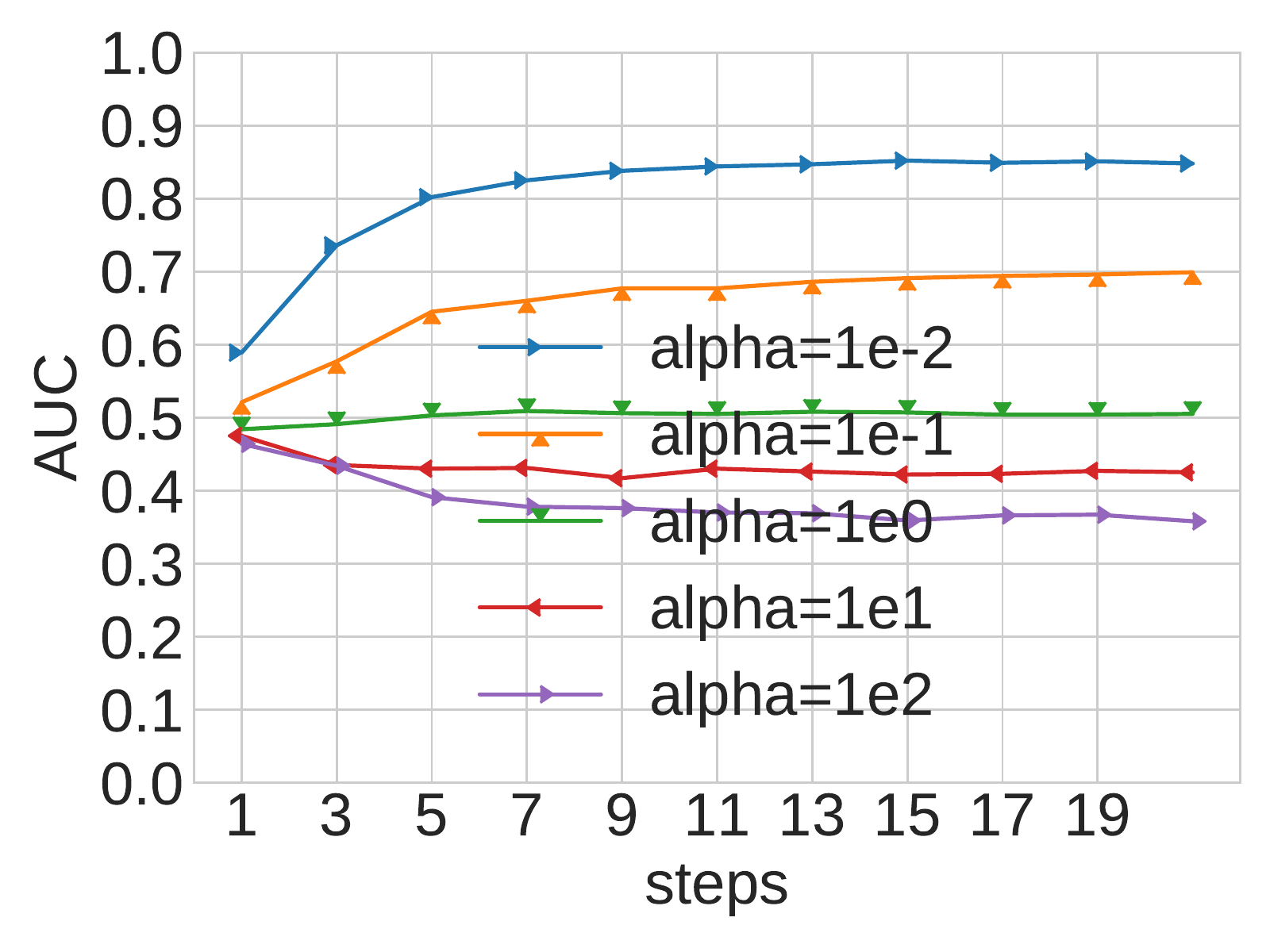}}
        \subfigure[acc. after attack]{\includegraphics[height=3.0cm]{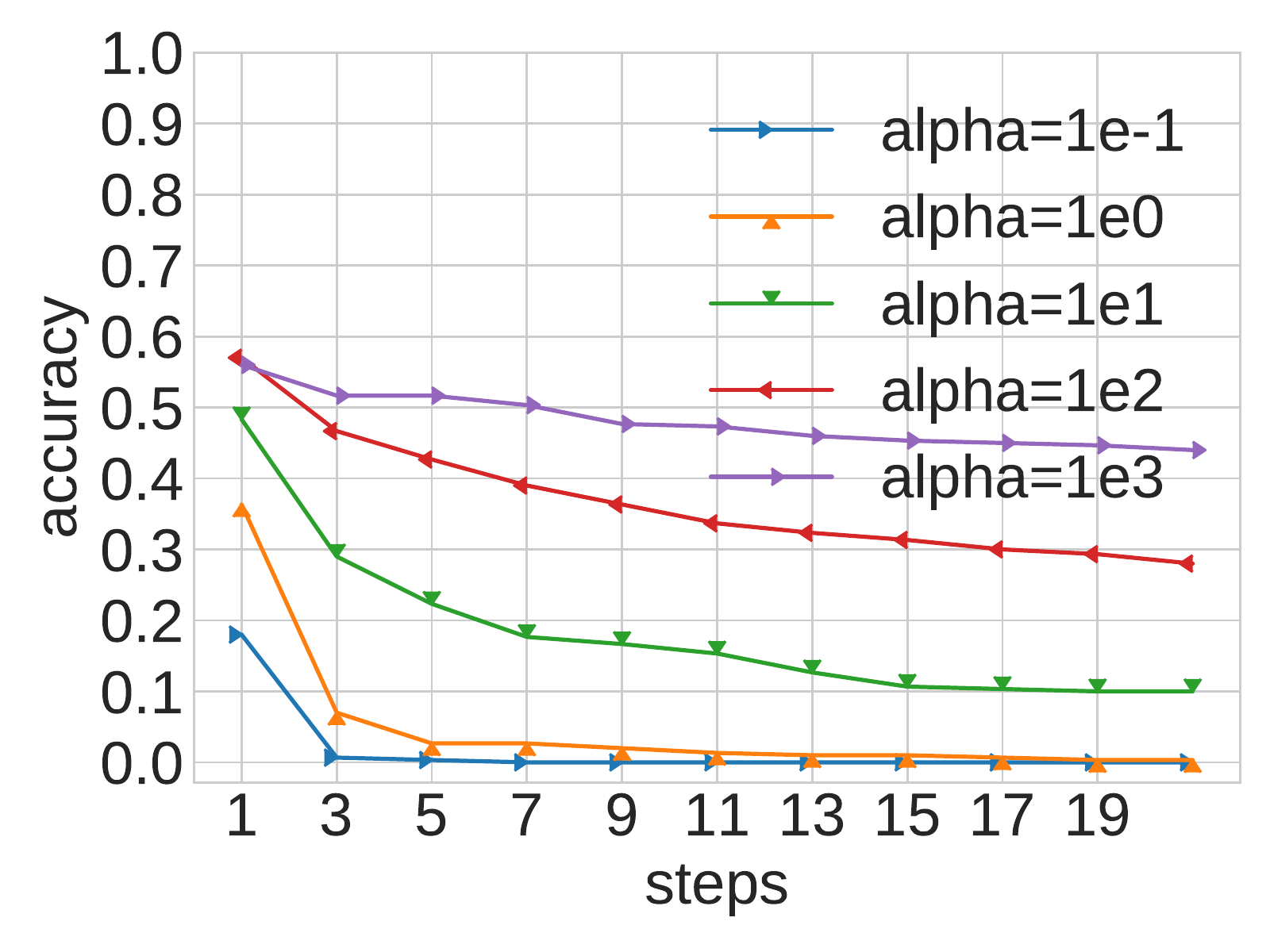}}
        \subfigure[AUC of detection]{\includegraphics[height=3.0cm]{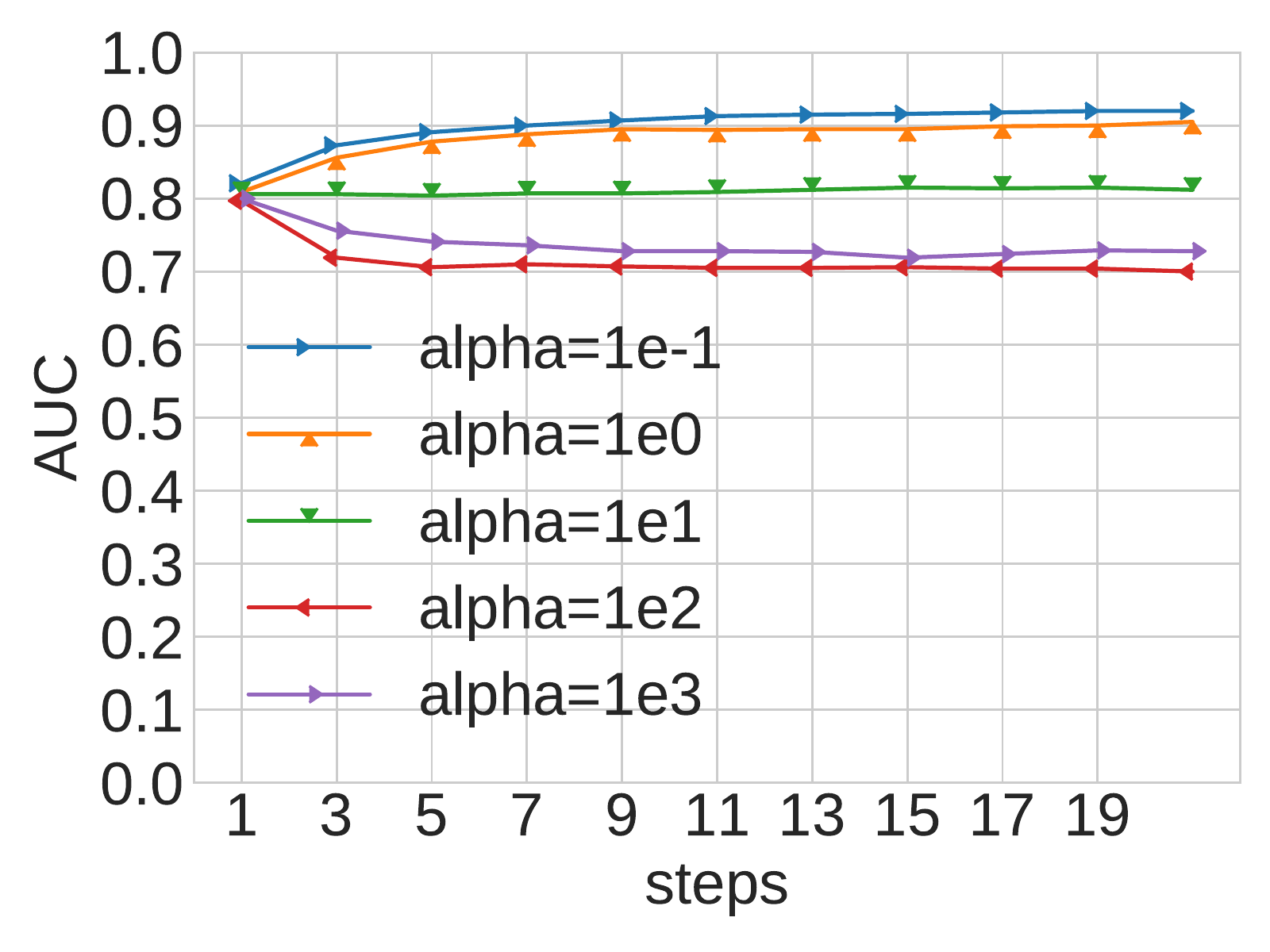}}
        \caption{
            Accuracy of victim model under the adaptive PGD attack with different $\alpha$. (a, b) MNIST, (c, d) SVHN.
        \label{Fig:adp}
    }\quad
\end{figure*}

We evaluate this adaptive attack strategy with the same setting used in Sec.~\ref{Sec:results}, and the results for MNIST and SVHN are depicted in Fig.~\ref{Fig:adp}. 
Observing Figs.~\ref{Fig:adp}(a) and (c), if the attacker set a larger value of $\alpha$ (e.g., $\alpha = 1e2$) and penalize the loss function $\textnormal{L}_{\textnormal{RE}}$ more (to avoid being detected), it is clear that the accuracy of the model is still high (i.e., the success rate of attack is low).
That said, it becomes harder for the attacker to succeed in attacking when the detector DRR is deployed.
Similarly, observing Figs.~\ref{Fig:adp}(b) and (d), if the attacker set a smaller value of $\alpha$ (e.g., $\alpha = 1e-2$) and focus more in attack, the AUC of the detector is still high (i.e., the successful attack can be still detected). 
It is worth mentioning similar results can be also observed on CIFAR10/CIFAR100. 
To conclude, compared to other self-supervised adversarial detectors, DRR is the first of its kind that offers good resistance to adaptive adversarial attacks. 


 \section{Detecting Textual Adversarial Examples}
DRR is based on the very basic fact that adversarial examples lead to wrong decisions of neural models without changing semantics of the examples, and this fact is data format-agnostic. We hereby exemplify the effectiveness of DRR on textual data. 

The works in \cite{Ren2019GeneratingNL,Alzantot2018GeneratingNL} propose to generate textual adversarial examples to attack natural language processing (NLP) models. 
These works try to fool victims models by modifying similar words or characters. 
For word-level attacks, attackers typically add perturbations by changing words to synonyms. 
For character-level attacks, attackers typically add perturbations by modifying characters to similar ones, for example, by replacing character 'o' by '0'. 
We still focus on the detection of textual adversarial examples of classification models.

\begin{figure}[h]
        \centering 
        \includegraphics[height=3.5cm]{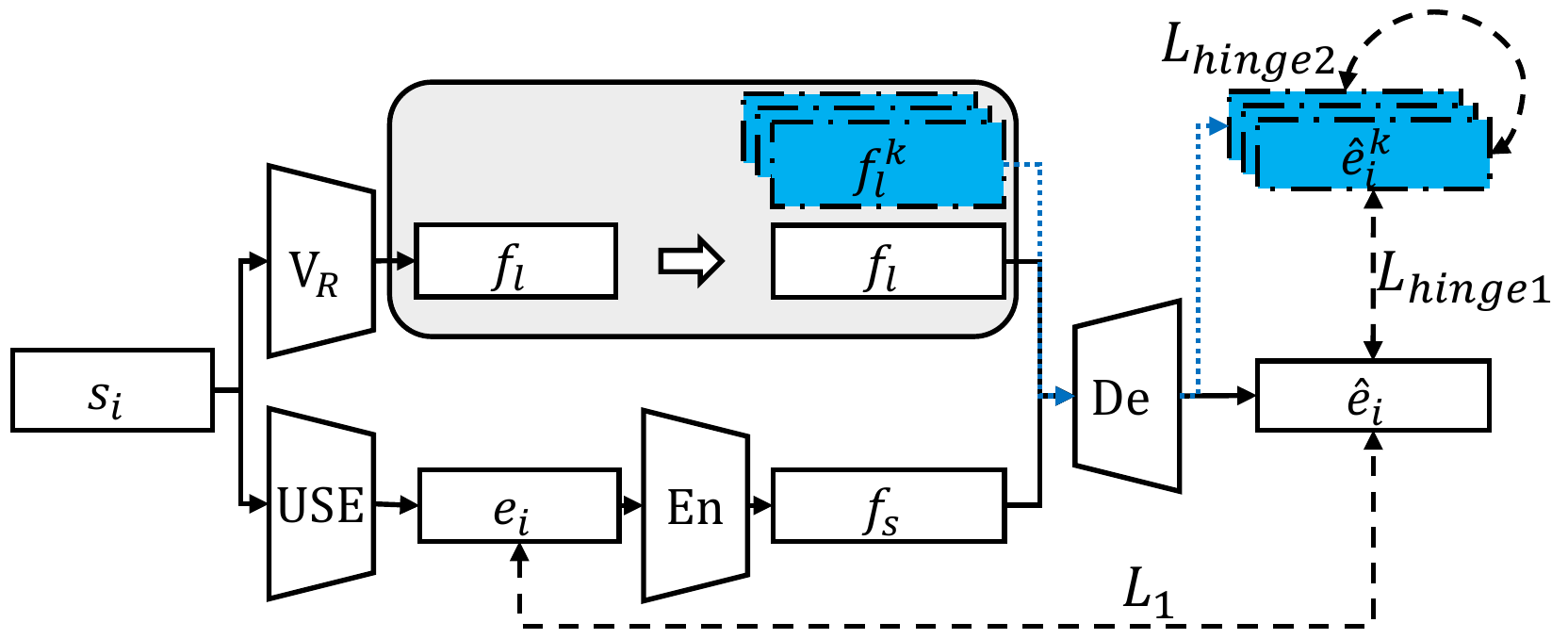}
        \caption{
            Overview of DRR in NLP.
        }
        \label{Fig:DRR_nlp}
    \vspace*{-10pt}
\end{figure}

Different from image data, natural language is discrete and harder to reconstruct. Instead of directly reconstructing natural language sentences, we choose to transform sentences to embeddings before encoding and decoding. Then we use the reconstruction error of embeddings as a metric to detect adversarial examples.
As shown in Fig.~\ref{Fig:DRR_nlp}, we first translate discrete textual data into continuous embedding by using the Universal Sentence Encoder for English ($\mathsf{USE}$) \cite{Cer2018UniversalSE}. $\mathsf{USE}$ is widely used for representing natural language, and the embedding produced by USE can preserve the information of natural language sentences.

Similar to the method mentioned in Sec.~\ref{Sec:train_drr}, the label feature $f_{l}'$ is obtained via
\begin{IEEEeqnarray}{rCl}
    y &=& \mathsf{V}(s_i) = \softmax(\mathsf{V}_z(s_i)) , \\
    f_{l}' &=& \mathsf{V}_R(s_i) = R_S(\mathsf{V}_z(s_i)),
    \label{Eq:textual1}
\end{IEEEeqnarray}
where $s_i$ represents the input sentences. We use the same rescale function $R_S$ as in Eq.~(\ref{Eq:rescale_func}). 
In contrast to image tasks, in textual tasks, the semantic feature $f_{s}'$ is encoded over $\mathsf{USE}$-generated embedding $e_i$ instead of the original input sentences. Similarly, the outputs of decoder are reconstructed embedding vectors. This process is characterized by 
\begin{IEEEeqnarray}{rCl}
    e_i &=& \mathsf{USE}(s_i), \\
    f_{s}' &=& \mathsf{En}(e_i), \\
    \hat{e}_i &=& \mathsf{De}(f_{l}', f_{s}').
    \label{Eq:textual2}
\end{IEEEeqnarray}
We then apply the idea of using disentangled feature to mimic both benign and adversarial examples for training DRR over embedded textual data, which is similar to that of Sec.~\ref{Sec:train_drr}. 

As shown in Fig.~\ref{Fig:DRR_nlp}, for benign embedding $e_i$, we use $\mathsf{MAE}$ to make sure it is similar to the its reconstructed version $\hat{e}_i$, thus the associated loss is: 
\begin{IEEEeqnarray}{rCl}
    \textnormal{L}_1' &=& \mathbb{E}_{e_i} \mathsf{MAE}(e_i, \hat{e}_i).
    \label{Eq:text_loss_reconstruction}
\end{IEEEeqnarray}
Moreover, we modify $\textnormal{L}_{hinge}$ in the same way as above to make them fit into textual attack. The textual version hinge loss functions are as follows: 
\begin{IEEEeqnarray}{rCl}
    \textnormal{L}_{hinge1}' &=& \mathbb{E}_{x_i} \left[ \max \Big(0, d-\min_{\substack{k_1, k_2 \in \Sigma,\\ k_1 \neq k_2}}\big( \mathsf{MAE}( \hat{e}^{k_1}_i, \hat{e}^{k_2}_i ) \big) \Big) \right],  \nonumber \\ \\
    \textnormal{L}_{hinge2}' &=& \mathbb{E}_{x_i} \left[ \max \Big(0, d-\min_{ \substack{k\in \Sigma,\\ k \neq i}} \big( \mathsf{MAE}( \hat{e}^k_i, e_i ) \big) \Big) \right], \\ 
    \textnormal{L}_{hinge}' &=& \textnormal{L}_{hinge1}+\textnormal{L}_{hinge2}, 
    \label{Eq:text_hinge_loss}
\end{IEEEeqnarray}
The final loss $\textnormal{Loss}_{text}$ to train encoder and decoder of DRR in textual tasks is thus
\begin{IEEEeqnarray}{rCl}
    \textnormal{Loss}_{text} = \lambda \textnormal{L}_{1}' + \textnormal{L}_{hinge}'.
\end{IEEEeqnarray}

To evaluate the performance of DRR for textual attacks, we choose the widely used BERT \cite{Devlin2019BERTPO} and ALBERT \cite{Lan2020ALBERTAL} as victim models, and train them with datasets SST-2 \cite{Socher2013RecursiveDM} and AG\_News \cite{Zhang2015CharacterlevelCN}, respectively. 
The SST-2 consists of 11,855 single sentences extracted from movie reviews, and there are two classes in SST-2: negative and positive.
The AG\_News contains more than one million news articles, which contains 4 kinds of news, including  World, Sports, Business and Sci/Tech.
BERT and ALBERT are pre-trained language models, after fine-tuned on SST-2 and AG\_News, they can be used as classification models.

We adopt two state-of-the-art word-level attacks, Probability Weighted Word Saliency (PWWS) \cite{Ren2019GeneratingNL} and Genetic \cite{Alzantot2018GeneratingNL}, to generate adversarial examples to evaluate the effectiveness of DRR. 
PWWS and Genetic generate adversarial examples by replacing synonym words. 
Follow the setting of previous sections, we still adopt AUC to evaluate the performances of DRR in detecting textual attack. From the AUC value of each attack for different datasets listed in Table~\ref{tab:auc_nlp}, it is concluded DRR is also valid for defending against adversarial textual examples.

\begin{table}[t]
    \caption{AUC of different textual attack over different datasets.}
    \label{tab:auc_nlp}%
    \centering
      \begin{tabular}{l|rr}
      \toprule
            & \multicolumn{1}{l}{SST-2} & \multicolumn{1}{l}{AG\_News} \\
      \midrule
      PWWS  & 0.79  & 0.821 \\
      Genetic & 0.78  & 0.888 \\
      \bottomrule
      \end{tabular}%
    \captionsetup{type=table}
\end{table}

\section{Conclusion}
\label{Sec:conclusion}
In this study, we propose to use disentangled representation for self-supervised adversarial examples detection. 
The proposed DRR is based on the very nature of adversarial examples: misleading classification results without changing the semantics of inputs (a lot). 
With disentangled label and semantic features, this nature inspires us to construct counterexamples to better guide the training of DRR. 
Compared with previous self-supervised detectors, DRR generally performs better under various measurements over different datasets and different adversarial attack methods. 
Not surprisingly, compared with other AE-based adversarial detectors, DRR is also more robust to adaptive adversaries. 
As exemplified over textual data, the rationale for designing DRR is universal, it is possible to extend DRR for defensing adversarial attacks in other domains. 
These merits make DRR a promising candidate, when combined with other proactive strategies, for the defense of adversarial attacks in real applications. 

\bibliographystyle{ieeetr}
\bibliography{ieee}

\end{document}